\documentclass[journal]{IEEEtran}
\usepackage{cite}
\usepackage{graphicx}
\usepackage[nice]{nicefrac}
\usepackage{amsmath}
\usepackage{amssymb}
\usepackage{amsthm}
\usepackage{booktabs}
\usepackage{url}
\usepackage{subfigure}
\usepackage{listings}
\usepackage{xspace}

\usepackage{algorithm}
\usepackage{algorithmicx}
\usepackage{algpseudocode}
\usepackage{color}
\makeatletter
\DeclareRobustCommand\onedot{\futurelet\@let@token\@onedot}
\def\@onedot{\ifx\@let@token.\else.\null\fi\xspace}
\def\eg{\emph{e.g}\onedot} 
\def\ie{\emph{i.e}\onedot} 
 
\def\etc{\emph{etc}\onedot}

\DeclareMathOperator*{\argmax}{arg\,max}
\makeatother

\begin{document}

\title{OpenHoldem: A Benchmark for Large-Scale\\Imperfect-Information Game Research}

\author{Kai Li,~\IEEEmembership{Member,~IEEE},
	    Hang Xu,
	    Enmin Zhao,
	    Zhe Wu,
	    and~Junliang Xing,~\IEEEmembership{Senior~Member,~IEEE}
\thanks{Kai Li, Hang Xu, and Enmin Zhao contributed equally to this work. Junliang Xing is the corresponding author.}
\thanks{Kai Li, Hang Xu, Enmin Zhao, Zhe Wu, and Junliang Xing are with the Institute of Automation, Chinese Academy of Sciences, and School of Artificial Intelligence, University of Chinese Academy of Sciences, Beijing, China (e-mail: kai.li@ia.ac.cn; xuhang2020@ia.ac.cn; zhaoenmin2018@ia.ac.cn; wuzhe2019@ia.ac.cn; jlxing@nlpr.ia.ac.cn).}
\thanks{This work was supported in part by the Natural Science Foundation of China under Grant No. 62076238 and 61902402, in part by the National Key Research and Development Program of China under Grant No. 2020AAA0103401, in part by the CCF-Tencent Open Fund, and in part by the Strategic Priority Research Program of Chinese Academy of Sciences under Grant No. XDA27000000.}
}

\markboth{Journal of \LaTeX\ Class Files,~Vol.~XX, No.~XX, XX~2021}%
{Shell \MakeLowercase{\textit{et al.}}: Bare Demo of IEEEtran.cls for IEEE Journals}

\maketitle

\begin{abstract}
Owning to the unremitting efforts by a few institutes, significant progress has recently been made in designing superhuman AIs in No-limit Texas Hold'em (NLTH), the primary testbed for large-scale imperfect-information game research. 
However, it remains challenging for new researchers to study this problem since there are no standard benchmarks for comparing with existing methods, which seriously hinders further developments in this research area.
In this work, we present OpenHoldem, an integrated toolkit for large-scale imperfect-information game research using NLTH. 
OpenHoldem makes three main contributions to this research direction: 1) a standardized evaluation protocol for thoroughly evaluating different NLTH AIs, 2) four publicly available strong baselines for NLTH AI, and 3) an online testing platform with easy-to-use APIs for public NLTH AI evaluation.
We have released OpenHoldem at \textit{\url{holdem.ia.ac.cn}}, hoping it facilitates further studies on the unsolved theoretical and computational issues in this area and cultivate crucial research problems like opponent modeling and human-computer interactive learning.

\end{abstract}


\IEEEpeerreviewmaketitle

\section{Introduction}
From its inception, artificial intelligence (AI) research has been focusing on building agents that can play games like humans. 
Both Turing~\cite{turing1953faster} and Shannon~\cite{shannon1950xxii} developed programs for playing chess to validate initial ideas in AI. 
For more than half a century, games have continued to be AI testbeds for novel ideas, and the resulting achievements have marked important milestones in the history of AI~\cite{schaeffer1997one,campbell2002deep,mnih2015human,silver2016mastering,silver2017mastering,silver2018general,schrittwieser2020mastering,jaderberg2019human,vinyals2019grandmaster,berner2019dota,li2020suphx,ye2020mastering,NEURIPS2020_06d5ae10,moravvcik2017deepstack,brown2018superhuman}. 
Notable examples include the checkers-playing bot Chinook winning a world championship against top humans~\cite{schaeffer1997one}, Deep Blue beating Kasparov in chess~\cite{campbell2002deep}, and AlphaGo defeating Lee Sedol~\cite{silver2016mastering} in the complex ancient Chinese game Go. 
Although substantial progress has been made in solving these large-scale perfect-information games that all players know the exact state of the game at every decision point, it remains challenging to solve large-scale imperfect-information games that require reasoning under the uncertainty about the opponents' hidden information. 
The hidden information is omnipresent in real-world strategic interactions, such as business, negotiation, and finance, making the research of imperfect-information games particularly important both theoretically and practically.

Poker has a long history as a challenging problem for developing algorithms that deal with hidden information~\cite{nash1951non,rubin2011computer}. 
The poker game involves all players being dealt with some private cards visible only to themselves, with players taking structured turns making bets, calling opponents' bets, or folding. 
As one of the most popular global card games, poker has played an essential role in developing general-purpose techniques for imperfect-information games. 
In particular, No-limit Texas Hold'em (NLTH), the world's most popular form of poker, has been the primary testbed for imperfect-information game research for decades because of its large-scale decision space and strategic complexity. 
For example, Heads-up No-limit Texas Hold'em (HUNL), the smallest variant of NLTH, has $10^{161}$ decision points~\cite{johanson2013measuring} which makes it almost impossible to solve directly.

There have been many efforts to design poker AIs for NLTH over the past few years~\cite{jackson2013slumbot,brown2015hierarchical}. 
Most of these systems exploit some equilibrium-finding algorithms, \eg, counterfactual regret minimization (CFR)~\cite{zinkevich2008regret}, with various abstraction strategies to merge similar game states to reduce the size of the game tree. 
Recently, a series of breakthroughs have been made in the NLTH AI research community. 
DeepStack~\cite{moravvcik2017deepstack}, which combines the continual re-solving and the depth-limited sparse look-ahead algorithms, defeated 10 out of 11 professional poker players by a statistically significant margin. 
Libratus~\cite{brown2018superhuman} defeated a team of four top HUNL-specialist professionals by using a nested safe subgame solving algorithm with an extensible blueprint strategy. 
Pluribus~\cite{brown2019superhuman} defeated elite human professional players in six-player NLTH by extending the techniques behind Libratus.

Although many important milestones have been achieved in NLTH AI research in recent years, the problem is far from being solved, and there remain many theoretical and computational issues to be addressed. 
For example, the game-theoretic solution for multiplayer NLTH, the best way to game tree abstraction, more efficient equilibrium-finding algorithms that converge faster and consume fewer resources, \etc. 
To solve these challenges, further studies are urgently needed. However, one main obstacle to further research in NLTH AI is the lack of standard benchmarks in this area. 
First, there are no standard evaluation protocols in this community; different papers use different evaluation metrics, making comparisons of different methods difficult. 
Second, there is no publicly available baseline AI which can serve as a starting point for future improvements. 
Third, there are no public easy-to-use platforms for researchers to test the performance of their AIs at any time.

Considering the important role of standard benchmarks in AI development, we present \textit{\textbf{OpenHoldem}}, a benchmark for NLTH AI research developed to boost the studies on large-scale imperfect-information games. 
OpenHoldem provides an integrated toolkit for evaluating NLTH AIs with three main components: the evaluation protocols, the baseline AIs, and a testing platform. For each component, we have made the following contributions to the community:
\begin{itemize}
	\item \textbf{For the evaluation part}, we propose to use four different evaluation metrics to test different algorithms from different aspects comprehensively.
	\item \textbf{For the baseline part}, we design and implement four different types of NLTH AIs: rule-based AI, CFR based static AI, DeepStack-like online AI, and deep reinforcement learning based AI. 
	These diverse AIs can serve as strong baselines for further development in this field.
	\item \textbf{For the platform part}, we develop an online testing platform with multiple NLTH AIs built-in. 
	Researchers can link their AIs to this platform through easy-to-use APIs to play against each other for mutual improvement.
\end{itemize}

Our proposed OpenHoldem provides a standardized benchmark for the NLTH AI research. 
The adopted approach, namely to propose an evaluation protocol via several metrics, the provision of baselines tested to have strong performances, and the establishment of an online testing platform, is perfectly rigorous and will allow algorithm improvements and comparisons with the state-of-the-arts, which impossible to do today without spending much time re-implementing other people's methods.
OpenHoldem can potentially have a significant impact on the poker AI research, and more generally in the AI community dealing with decision-making problems under uncertainty.
We hope that OpenHoldem makes the NLTH AI research easier and more accessible, and further facilitates the research of the key problems in large-scale imperfect-information games, such as large-scale equilibrium-finding, opponent modeling, human-computer interactive learning, and online exploiting sub-optimal opponents.

\section{Related Work}
Standard benchmarks have played an indispensable role in promoting the research in many AI tasks like speech recognition, computer vision, and natural language processing. 
For example, in the task of speech to text, the NIST Switchboard  benchmark~\cite{godfrey1992switchboard} helps reduce the word error rate from $19.3\%$ in 2000 to $5.5\%$ in 2017; 
In the task of image classification, the creation of the ImageNet~\cite{deng2009imagenet} benchmark has helped in the development of highly efficient models which reduce the image classification error rate from $26.2\%$ down to $1.8\%$; 
In the task of machine translation, the WMT benchmark helps the machine translation system achieves human-level performance on the Chinese to English translation task~\cite{hassan2018achieving}.
These benchmarks that have greatly influenced the research communities have some common characteristics: clear evaluation metrics, rich baseline models, and convenient online testing platforms. 
Motivated by this, we propose the OpenHoldem benchmark that meets the above requirements to facilitate the future development of general-purpose techniques for large-scale imperfect-information games.

There are already some benchmarks on game AI. 
Examples include the Atari environments in OpenAI Gym~\cite{brockman2016openai}, ViZDoom~\cite{wydmuch2018vizdoom}, and MineRL~\cite{guss2019minerl}, but most of these benchmarks are oriented towards the research of reinforcement learning algorithms. 
Recently, some benchmarks for game theory research have been proposed. 
For example, Google DeepMind releases the OpenSpiel~\cite{lanctot2019openspiel} benchmark, which contains a collection of environments and algorithms for research in n-player zero-sum and general-sum games. 
Although OpenSpiel implements many different kinds of games and state-of-the-art algorithms, it currently does not provide high-performance NLTH AIs.
RLCard~\cite{zha2019rlcard} developed by the Texas A\&M University includes many large-scale complex card games, such as Dou dizhu, Mahjong, UNO, Sheng Ji, and NLTH. 
However, most of the implemented baseline AIs are relatively weak. 
In contrast, the proposed OpenHoldem contains very strong baseline AIs, which can serve as a better starting point for future improvements.

Texas Hold'em, the primary testbed for imperfect information game research, has been studied in the computer poker community for years~\cite{rubin2011computer}. 
The earliest Texas Hold'em AIs are rule-based systems that consist of a collection of if-then rules written by human experts. 
For example, the early agents (\eg, Loki~\cite{billings1998opponent}) produced by the University of Alberta are mostly based on carefully designed rules. 
While the rule-based approach provides a simple framework for implementing Texas Hold'em AIs, the resulting handcrafted strategies are easily exploitable by observant opponents. 
Since 2006, the Annual Computer Poker Competition (ACPC)~\cite{bard2013annual} has greatly facilitated poker AI development, and many game-theoretic Texas Hold'em AIs are proposed~\cite{jackson2013slumbot,brown2015hierarchical}. 
These systems first use various abstraction strategies~\cite{johanson2013evaluating,ganzfried2014potential} to merge similar game states to reduce the game size, then exploit some equilibrium-finding algorithms (\eg, CFR~\cite{zinkevich2008regret} and its various variants~\cite{lanctot2009monte,tammelin2014solving,jackson2016compact,schmid2019variance}) to find the approximate Nash equilibrium strategies which are robust to different opponents.

Recently, the research on these game-theoretic approaches has made significant breakthroughs. 
Examples include DeepStack~\cite{moravvcik2017deepstack} proposed by the University of Alberta that defeats professional poker players by a large margin, Libratus~\cite{brown2018superhuman} from the Carnegie Mellon University that decisively defeats four top HUNL-specialist professionals, and Pluribus~\cite{brown2019superhuman} as a direct descendant of Libratus that defeats elite human professional players in six-player NLTH. 
Nevertheless, almost all of these Texas Hold'em AIs are not publicly available, making it very challenging for new researchers to study this problem further. 
Our OpenHoldem is the first open benchmark with publicly available strong baseline AIs for large-scale imperfect-information game research.

\section{Preliminaries}
Here we present some background knowledge needed for the rest of the paper. 
We first provide some notations to formulate imperfect-information games.
Next, we discuss the CFR algorithm which is the most commonly used equilibrium-finding algorithm for imperfect-information games. 
Finally, we introduce the game rule of no-limit Texas Hold'em. 

\subsection{Imperfect-Information Games}
Imperfect-information games are usually described by a tree-based formalism called \textbf{extensive-form games}~\cite{osborne1994course}. 
In an imperfect-information extensive-form game $\mathcal{G}$ there is a finite set $\mathcal{N}=\{1,\!\ldots\!,N\}$ of \textbf{players}, and there is also a special player $c$ called \textbf{chance};
$\mathcal{H}$ refers to a finite set of histories, each member $h\in \mathcal{H}$ denotes a possible \textbf{history} (or \textbf{state}), which consists of actions taken by players including chance; 
$g \sqsubseteq h$ denotes the fact that $g$ is equal to or a \textbf{prefix} of $h$;
$\mathcal{Z} \subseteq \mathcal{H}$ denotes the \textbf{terminal states} and any member $z\!\in\!\mathcal{Z}$ is not a prefix of any other states; 
$\mathcal{A}(h)=\{a:ha\!\in\!\mathcal{H}\}$ is the set of available \textbf{actions} in the non-terminal state $h\!\in\!\mathcal{H}\setminus\mathcal{Z}$; 
A \textbf{player function} $\mathcal{P}:\mathcal{H}\setminus\mathcal{Z}\rightarrow\mathcal{N}\cup\{c\}$ assigns a member of $\mathcal{N}\cup\{c\}$ to each non-terminal state in $\mathcal{H}\setminus\mathcal{Z}$, \ie, $\mathcal{P}(h)$ is the player who takes an action in state $h$.

For a state set $\{h \in \mathcal{H}: \mathcal{P}(h)=i\}$, $\mathcal{I}_{i}$ denotes an \textbf{information partition} of player $i$; 
A set $I_{i} \in \mathcal{I}_{i}$ is an \textbf{information set} of player $i$ and $I(h)$ represents the information set which contains the state $h$. 
If $g$ and $h$ belong to the same information set $I_{i}$, then the player $i$ cannot distinguish between them, so we can define $\mathcal{A}(I_{i}) = \mathcal{A}(h)$ and $\mathcal{P}(I_{i}) = \mathcal{P}(h)$ for arbitrary $h \in I_{i}$. 
We define $\vert\mathcal{I} \vert= \max_{i\in\mathcal{N}}\vert\mathcal{I}_i\vert$ and $\vert\mathcal{A}\vert = \max_{i\in\mathcal{N}}\max_{I_i\in\mathcal{I}_i}\vert \mathcal{A}(I_i) \vert$.
For each player $i \in \mathcal{N}$, a \textbf{utility function} $u_{i}(z)$ define the payoff received by player $i$ upon reaching a terminal state $z$.
$\Delta_i$ is the \textbf{range of payoffs} reachable by player $i$, \ie, $\Delta_i = \max_{z\in \mathcal{Z}}u_i(z) - \min_{z\in \mathcal{Z}}u_i(z)$ and $\Delta=\max_{i\in\mathcal{N}}\Delta_i$.

A \textbf{strategy profile} $\sigma=\{\sigma_{i}| \sigma_{i} \in \Sigma_{i}, i \in \mathcal{N}\}$ is a specification of strategies for all players, where $\Sigma_{i}$ is the set of all possible strategies for player $i$, and $\sigma_{-i}$ refers to the strategies of all players other than player $i$. 
For each player $i \in \mathcal{N}$, its strategy $\sigma_{i}$ assigns a distribution over $\mathcal{A}(I_{i})$ to each information set $I_{i}$ of player $i$. 
The strategy of the chance player $\sigma_{c}$ is usually a fixed probability distribution.
$\sigma_{i}(a|h)$ denotes the probability of action $a$ taken by player $i \in \mathcal{N}$ at state $h$. 
In imperfect information games, $\forall h_{1}, h_{2} \in I_{i}$, we have $\sigma_{i}(I_{i})=\sigma_{i}(h_{1})=\sigma_{i}(h_{2})$. 
The \textbf{state reach probability} of $h$ is denoted by $\pi^{\sigma}(h)$ if all players take actions according to the strategy profile $\sigma$. 
The state reach probability can be composed into each player's contribution, \ie, $\pi^{\sigma}(h)=\prod_{i \in \mathcal{N} \cup \{c\}}\pi^{\sigma}_{i}(h)=\pi^{\sigma}_{i}(h) \pi^{\sigma}_{-i}(h)$, where $\pi^{\sigma}_{i}(h)=\prod_{h'a \sqsubseteq h, \mathcal{P}(h')=i }\sigma_{i}(a|h')$ is player $i's$ contribution and $\pi^{\sigma}_{-i}(h)=\prod_{h'a \sqsubseteq h, \mathcal{P}(h') \neq i }\sigma_{\mathcal{P}(h')}(a|h')$ is all players' contribution except player $i$. 
The \textbf{information set reach probability} of $I_{i}$ is defined as $\pi^{\sigma}(I_{i})=\sum_{h \in I_{i}}\pi^{\sigma}(h)$. 
The \textbf{interval state reach probability} from state $h'$ to $h$ is defined as $\pi^{\sigma}(h', h) = \pi^{\sigma}(h)/\pi^{\sigma}(h')$ if $h' \sqsubseteq h$. 
$\pi^{\sigma}_{i}(I_{i})$, $\pi^{\sigma}_{-i}(I_{i})$, $\pi_{i}^{\sigma}(h', h)$, and $\pi_{-i}^{\sigma}(h', h)$ are defined similarly.

For each player $i \in \mathcal{N}$, the \textbf{expected utility} $u^{\sigma}_{i}=\sum_{z \in \mathcal{Z}}\pi^{\sigma}(z) u_{i}(z)$ under a strategy profile $\sigma$ is the expected payoff of player $i$ obtained at all possible terminal states. 
The \textbf{best response} to the strategy profile $\sigma_{-i}$ is any strategy $\sigma^{*}_{i}$ of player $i$ that achieves optimal payoff against $\sigma_{-i}$, \ie, $\sigma^{*}_{i}=\argmax_{\sigma_{i}' \in \Sigma_{i}}{u^{(\sigma_{i}', \sigma_{-i})}_{i}}$. 
For the two-player zero-sum games, \ie, $\mathcal{N}=\{1,2\}$ and $\forall z \in \mathcal{Z}, u_{1}(z)+u_{2}(z)=0$, 
the \textbf{Nash equilibrium} is the most commonly used solution concept which is a strategy profile $\sigma^{*}=(\sigma^{*}_{1}, \sigma^{*}_{2})$ such that each player's strategy is the best response to the other. 
An \textbf{$\epsilon$-Nash equilibrium} is an approximate Nash equilibrium, whose strategy profile $\sigma$ satisfies: $\forall i \in \mathcal{N}$, $u_{i}^{\sigma} + \epsilon \geq \max_{\sigma_{i}' \in \Sigma_{i}}{u^{(\sigma_{i}', \sigma_{-i})}_{i}}$. 
The \textbf{exploitability} of a strategy $\sigma_{i}$ is defined as $\epsilon_{i}(\sigma_{i}) = u_{i}^{\sigma^{*}} - u_{i}^{(\sigma_{i}, \sigma^{*}_{-i})}$. 
A strategy is unexploitable if $\epsilon_{i}(\sigma_{i})=0$.

\subsection{Counterfactual Regret Minimization}
Counterfactual Regret Minimization (CFR)~\cite{zinkevich2008regret} is an iterative algorithm for computing approximate Nash equilibrium in imperfect-information games and is widely used in NLTH AI.
CFR frequently uses \textbf{counterfactual value}, which is the expected payoff of an information set given that player $i$ tries to reach it.
Formally, for player $i$ at an information set $I\in\mathcal{I}_i$ given a strategy profile $\sigma$, the counterfactual value of $I$ is $v_i^\sigma(I)=\sum_{h\in I}(\pi_{-i}^\sigma(h)\sum_{z\in \mathcal{Z}}(\pi^\sigma(h, z)u_i(z))$.
The counterfactual value of an action $a$ in $I$ is $v_i^\sigma(a|I)=\sum_{h\in I}(\pi_{-i}^\sigma(h)\sum_{z\in \mathcal{Z}}(\pi^\sigma(ha, z)u_i(z))$.

CFR typically starts with a random strategy $\sigma^1$. 
On each iteration $T$, CFR first recursively traverses the game tree using the strategy $\sigma^T$ to calculate the \textbf{instantaneous regret} $r_i^T(a|I)$ of not choosing action $a$ in an information set $I$ for player $i$, \ie, $r_i^T(a|T) = v_i^{\sigma^T}(a|I) - v_i^{\sigma^T}(I)$.
Then CFR accumulates the instantaneous regret to obtain the \textbf{cumulative regret} $R_i^T(a|I) = \sum_{t=1}^T r_i^t(a|I)$ and uses regret-matching~\cite{blackwell1956analog} to calculate the new strategy for the next iteration:
\begin{equation}
	\resizebox{0.8\linewidth}{!}
	{
		$\sigma_i^{T+1}(a|I)=\left\{\begin{array}{cl}
			\frac{R_i^{T,+}(a|I)}{\sum_{a^{\prime} \in \mathcal{A}(I)} R_{i}^{T,+}\left( a^{\prime}| I \right)}, & \sum_{a^{\prime}} R_i^{T,+}\left( a^{\prime}| I \right)>0 \\
			\frac{1}{|\mathcal{A}(I)|}, & \text { otherwise }
		\end{array}\right.$
	}
\end{equation}
where $R_i^{T, +} (a|I) = \max(R_i^T(a|I), 0)$.

In two-player zero-sum imperfect-information games, if both players play according to CFR on each iteration then their \textbf{average strategies} $\bar{\sigma}^T$ converge to an $\epsilon$-Nash equilibrium in\ $\mathcal{O}(|\mathcal{I}|^2|\mathcal{A}|\Delta^2/\epsilon^2)$ iterations~\cite{zinkevich2008regret}.
$\bar{\sigma}^T$ is calculated as:
\begin{equation}
	\scriptsize
	\begin{aligned}
		\!\!\!\!\!S_i^{T}(a|I)\!\!=\!\!\sum_{t=1}^{T}\left(\pi_{i}^{\sigma^{t}}(I) \sigma_{i}^{t}(a|I)\right), \bar{\sigma}_i^T(a|I )\!\!=\!\!\frac{S_i^T(a|I)}{\sum_{a'\in\mathcal{A}(I)}S_i^T( a'|T)}.
		\label{eq:ave_strategy}
	\end{aligned}
\end{equation}
Thus, CFR is a ready-to-use equilibrium finding algorithm in two-player zero-sum games.

\subsection{No-limit Texas Hold'em}
No-limit Texas hold'em (NLTH) has been the most widely played type of poker for more than a decade. 
The heads-up (\ie, two-player) variant prevents opponent collusion and allows a clear winner to be determined, so heads-up no-limit Texas hold'em (HUNL) becomes the primary testbed in the computer poker and game theory communities.
HUNL is a repeated game in which the two players play a match of individual games, usually called \textbf{hands}.
On each hand, one player will win some number of \textbf{chips} from the other player, and the goal is to win as many chips as possible throughout the match.
In this paper, we follow the standard form of HUNL poker agreed upon by the research community~\cite{bard2013annual}, where each player starts each hand with a \textbf{stack} of \$20,000 chips.
Resetting the stacks after each hand allows for each hand to be an independent sample of the same game and is called ``Doyle's Game'', named for the professional poker player Doyle Brunson who publicized this variant.

HUNL consists of four rounds of betting.
On each round of betting, each player can choose to either \textbf{fold}, \textbf{call}, or \textbf{raise}.
If a player folds, the game will end with no player revealing their private
cards, and the opponent will take the \textbf{pot}.
If a player calls, he or she places several chips in the pot by matching the amount of chips entered by the opponent.
If a player raises by $x$, he or she adds $x$ more chips to the pot than the opponent.
A raise of all remaining chips is called an \textbf{all in} bet.
A betting round ends if each player has taken actions and has entered the same amount of chips in the pot as every other player still in the hand.
At the beginning of a round, when there are no opponent chips yet to match, the raise action is called \textbf{bet}, and the call action is called \textbf{check}.
If either player chooses to raise first in a round, they must raise a minimum of \$100 chips.
If a player raises after another player has raised, that raise must be greater than or equal to the last raise.
The maximum amount for a bet or raise is the remainder of that player's stack, which is \$20,000 at the beginning of a hand.

\begin{figure}[t]
	\centering
	\includegraphics[width=0.95\linewidth]{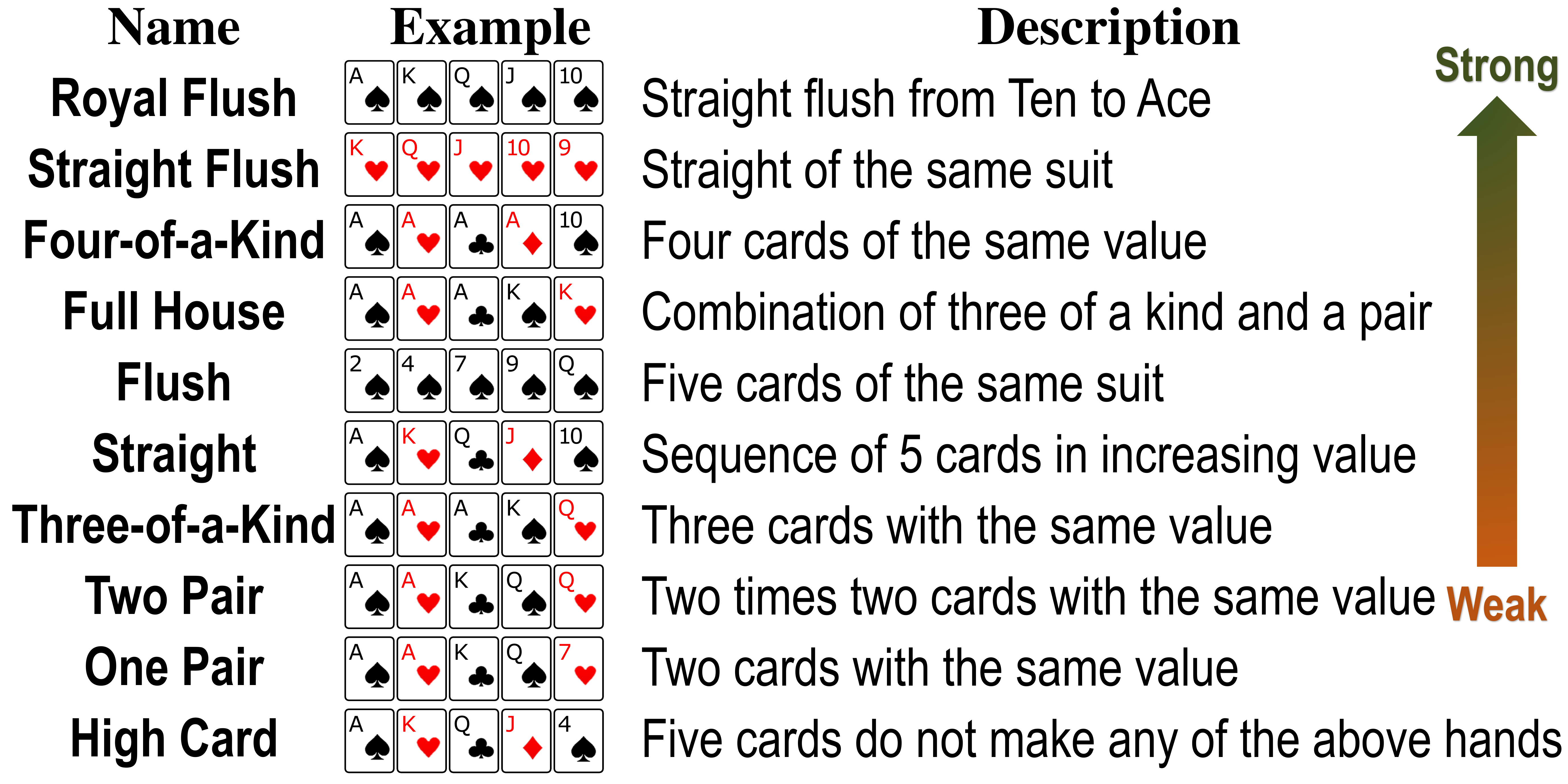}
	\caption{The hand strength of Texas hold'em poker.}
	\label{hand_strength}
\end{figure}

\begin{figure}[t]
	\centering
	\includegraphics[width=0.95\linewidth]{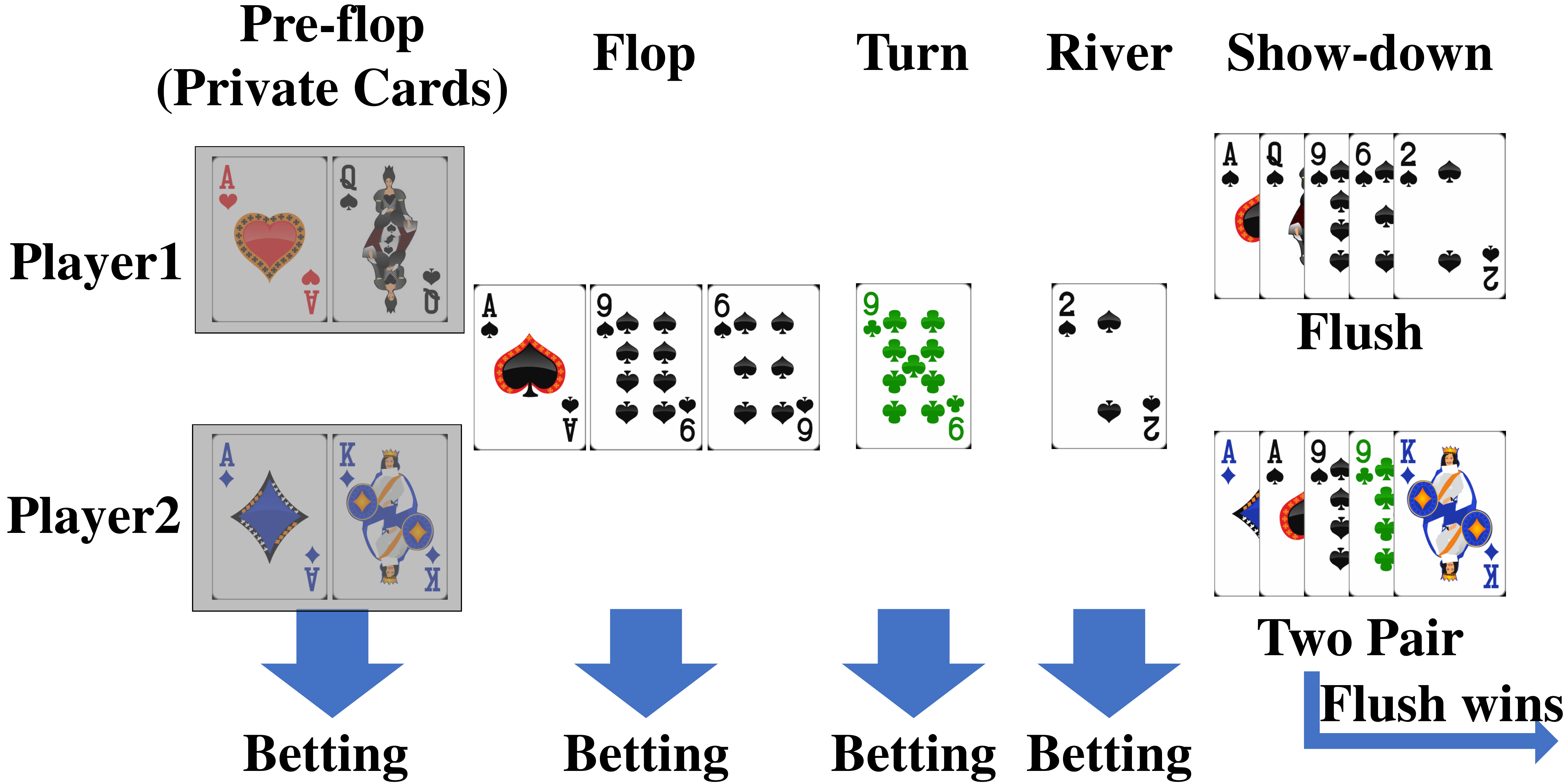}
	\caption{A visualized example of the four rounds in one HUNL game.}
	\label{holdem_rounds}
\end{figure}

In HUNL, at the beginning of each hand, the first player, \ie, P1, enters a \textbf{big blind} (usually \$100) into the pot; 
the second player, \ie, P2, enters a \textbf{small blind} which is generally half the size of the big blind; 
and both players are then dealt with two \textbf{hole (private) cards} from a standard 52-card deck.
There is then the first round of betting (called the \textbf{pre-flop}), where the second player P2 acts first.
The players alternate in choosing to fold, call or raise.
After the pre-flop, three \textbf{community (public) cards} are dealt face up for all players to observe, and the first player P1 now starts a similar round of betting (called the \textbf{flop}) to the first round.
After the flop round ends, another community card is dealt face up, and the third round of betting (called the \textbf{turn}) commences where P1 acts first.
Finally, a fifth community card is dealt face up, and a fourth betting round (called the \textbf{river}) occurs, again with P1 acting first.
If none of the players folds at the end of the fourth round, the game enters a \textbf{show-down} process: the private cards are revealed, the player with the best five-card poker hand (see Figure~\ref{hand_strength} for the hand strength), constructed from the player's two private cards and the five community cards, wins the pot.
In the case of a tie, the pot is split equally among the players.
For a better understanding of these rounds, Figure~\ref{holdem_rounds} provides a visualized example of the four rounds in one HUNL game.
A \textbf{match} consists of a large number of poker hands, in which the players alternate their positions as the first and the second player.
The rules of Six-player NLTH and HUNL are roughly the same. For the detailed rules of Six-player NLTH, please refer to the supplementary materials of~\cite{brown2019superhuman}.

Since NLTH can be played for different stakes, such as a big blind being worth \$0.01 or \$1000, it is inappropriate to measure the performance by chips, so players commonly measure their performance over a match as their average number of big blinds won per hand.
The computer poker community has standardized on the unit \textbf{milli-big-blinds per hand}, or mbb/h, where one milli-big-blind is one thousandth of one big blind.
For example, a player that always folds will lose 750 mbb/h (by losing 1000 mbb as the big blind and 500 as the small blind).

\section{OpenHoldem}
\begin{figure}[t]
	\centering
	\includegraphics[width=0.9\linewidth]{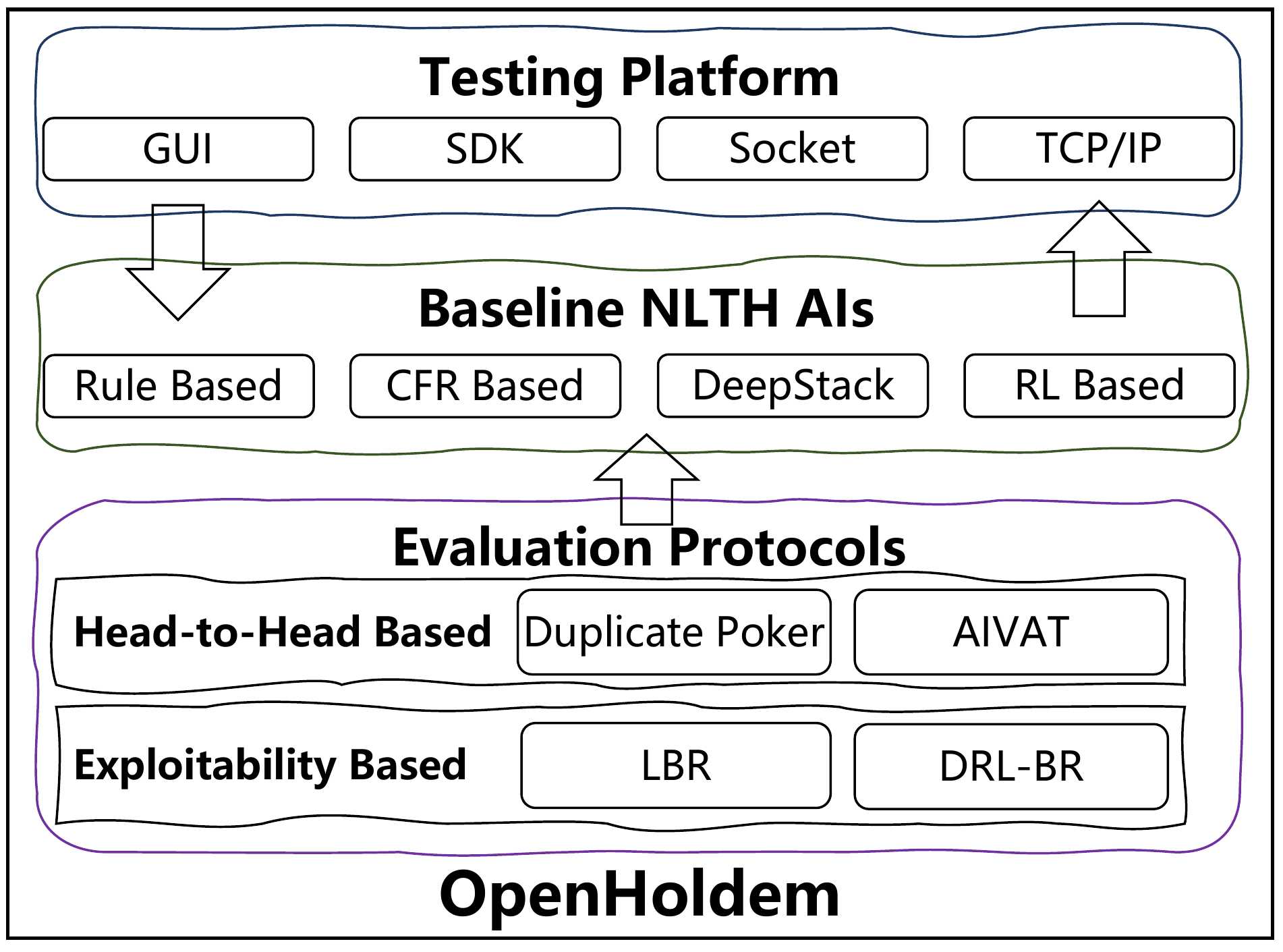}
	\caption{OpenHoldem provides an integrated toolkit for large-scale imperfect-information game research using NLTH with three main components: the evaluation protocols, the baseline NLTH AIs, and an online testing platform.}
	\label{OpenHoldem}
\end{figure}

As shown in Figure~\ref{OpenHoldem}, the proposed OpenHoldem benchmark for large-scale imperfect information game research consists of three parts: the evaluation protocols, the baseline AIs, and an online testing platform. 
Next, we will expatiate these three parts respectively.

\subsection{Evaluation Protocols}
Evaluating the performance of different NLTH agents is challenging due to the inherent variance present in the game. 
A better agent may lose in a short period simply because it was dealt with weaker cards. 
Moreover, different papers use different evaluation metrics, making comparisons of different methods difficult. 
In OpenHoldem, we propose using the following evaluation metrics to test different algorithms from different aspects thoroughly.

\subsubsection{Head-to-Head Based Evaluation Metrics}
One of the main goals of agent evaluation is to estimate the expected utility $u^{\sigma}_{i}$ of some player $i \in \mathcal{N}$ given a strategy profile $\sigma$. 
If the game is small, one can compute this expectation exactly by enumerating all terminal states, \ie, $u^{\sigma}_{i}=\sum_{z \in \mathcal{Z}}\pi^{\sigma}(z) u_{i}(z)$. 
In the large-scale NLTH, however, this approach is unpractical. 
The most commonly used approach to approximately estimate $u^{\sigma}_{i}$ is sampling. 
Specifically, the NLTH agents repeatedly play against each other, drawing independent samples $z_1,\ldots,z_T$ with the probability $\pi^{\sigma}(z)$. 
The estimator ${\hat{u}}^{\sigma}_{i}$ is simply the average utility,
\begin{equation}
	\label{MonteCarlo}
	{\hat{u}}^{\sigma}_{i} = \frac{1}{T}  \sum_{t=1}^{T} u_{i}(z_t).
\end{equation}
This estimator is unbiased, \ie, $E[{\hat{u}}^{\sigma}_{i}]=u^{\sigma}_{i}$, so the mean-squared-error (MSE) of ${\hat{u}}^{\sigma}_{i}$ is its variance,
\begin{equation}
	\text{MSE}({\hat{u}}^{\sigma}_{i})=\text{Var}[{\hat{u}}^{\sigma}_{i}] = \frac{1}{T} \text{Var}[u_{i}(z)].
\end{equation}
This sampling based approach is effective when the domain has little stochasticity, \ie, $\text{Var}[u_{i}(z)]$ is small, but this is not the case in NLTH. 
To alleviate the effects of randomness and ensure statistically significant results, we propose to use the following two variance reduction techniques in head-to-head based evaluation.

\textbf{Duplicate Poker} is a simple variance reduction technique that attempts to mitigate the effects of luck and is widely used in the Annual Computer Poker Competitions (ACPC)~\cite{bard2013annual}. 
For example, in HUNL, let us say agent $\mathcal{A}$ plays one seat and agent $\mathcal{B}$ plays the other seat. 
First, we let $\mathcal{A}$ and $\mathcal{B}$ play $M$ hands of poker, then we switch their seats and play another $M$ hands of poker with the same set of cards for each seat. 
By doing so, if agent $\mathcal{A}$ is dealt two aces in the first hand, then agent $\mathcal{B}$ will be dealt two aces in the $M+1$-th hand, so the effects of luck are significantly alleviated. 
The process of duplicate poker for multiplayer NLTH is similar.

\textbf{AIVAT} is a more principled variance reduction technique for evaluating performance of agents in imperfect-information games~\cite{burch2018aivat}. 
The core idea of AIVAT is to derive a real-valued function $\tilde{u}_{i}$ that is used in place of the true utility function $u_i$. 
On one hand, the expectation of $\tilde{u}_{i}(z)$ matches that of ${u}_{i}(z)$ for any choice of strategy profile $\sigma$, so ${\tilde{u}}^{\sigma}_{i} = \frac{1}{T}  \sum_{t=1}^{T} \tilde{u}_{i}(z_t)$ is also an unbiased estimator of the expected utility $u^{\sigma}_{i}$. 
On the other hand, the variance of $\tilde{u}_{i}(z)$ is designed to be smaller than that of $u_{i}(z)$, so $\text{MSE}({\tilde{u}}^{\sigma}_{i}) < \text{MSE}({\hat{u}}^{\sigma}_{i})$, \ie, ${\tilde{u}}^{\sigma}_{i}$ is a better estimator than ${\hat{u}}^{\sigma}_{i}$. 
More specifically, AIVAT adds a carefully designed control variate term for both chance actions and actions of players with known strategies, resulting in a provably unbiased low-variance evaluation tool for imperfect-information games. 
It is worth noting that duplicate poker and AIVAT can be combined to further reduce the variance.

\subsubsection{Exploitability Based Evaluation Metrics}
Most works on computer poker are to approximate a Nash equilibrium, \ie, produce a low-exploitability strategy. 
However, head-to-head evaluation has been shown to be a poor equilibrium approximation quality estimator in imperfect-information games~\cite{moravvcik2017deepstack}. 
For example, in the toy game of Rock-Paper-Scissors, consider the exact Nash equilibrium strategy (\ie, playing each option with equal probability) playing against a dummy strategy that always plays ``rock''. 
The head-to-head based evaluation results are a tie in this example, but the two strategies are vastly different in terms of exploitability. 
Therefore, the exploitability is also a crucial evaluation metric in imperfect-information games. 
The exploitability of one strategy can be measured by calculating its best-response strategy, but the large size of NLTH's game tree makes an explicit best-response computation intractable. 
We propose to use the following two techniques to calculate the exploitability approximately.

\textbf{Local Best Response} (LBR) is a simple and computationally inexpensive method to find a \emph{lower-bound} on a strategy's exploitability~\cite{lisy2017eqilibrium}. 
The most important concept in this algorithm is the agent's range, \ie, the probability distribution on each of the possible private cards the agent holds. 
Suppose we want to find the LBR of the agent $\mathcal{A}$ with known strategy $\sigma_a$. 
At the beginning of each poker hand, it is equally likely that $\mathcal{A}$ holds any pair of private cards. 
The probabilities of actions performed by  $\mathcal{A}$ depend on the private cards it holds. 
Knowing the strategy of $\mathcal{A}$, we can use
Bayes' theorem to infer the probabilities that $\mathcal{A}$ holds each of the private cards. 
Based on the range of $\mathcal{A}$, LBR greedily approximates the best response actions, \ie, the actions which maximize the expected utility under the assumption that the game will be checked/called until the end. 
Thus, LBR best-responds locally to the opponent's actions by looking only at one action ahead, providing a lower bound on the opponent's exploitability. 
LBR also relies on playing standard poker hands, so the variance reduction techniques (\eg, AIVAT) can be exploited to reduce the number of hands required to produce statistically significant results.

\textbf{Deep Reinforcement Learning Based Best Response (DRL-BR).} Because the game tree of NLTH is too large, the LBR algorithm does not explicitly compute a best-response strategy but uses its local approximation to play against the evaluated agent $\mathcal{A}$ directly. 
In DRL-BR, we try to explicitly approximate the best response strategy by training an DRL agent $\mathcal{B}$ against $\mathcal{A}$. 
More specifically, by treating $\mathcal{A}$ as part of the environment, then from the perspective of $\mathcal{B}$, the environment can be modeled as a Markov Decision Process (MDP).
$\mathcal{B}$ can leverage some suitable DRL algorithms (\eg, DQN~\cite{mnih2015human}, PPO~\cite{schulman2017proximal}, \etc) to learn to maximize its payoff from its experience of interacting with the environment, \ie, playing against $\mathcal{A}$. 
This approach turns the problem of finding the best response strategy into a single agent RL problem. 
An approximate solution of the MDP by RL yields an approximate best response to the evaluated agent $\mathcal{A}$. 
After obtaining the approximate best response $\mathcal{B}$, the head-to-head evaluation result (\eg, AIVAT) can be used to approximate the exploitability of $\mathcal{A}$ by having them repeatedly play against each other.

\subsection{Baseline AIs}
Despite significant progress in designing NLTH AIs in recent years, almost all of these AIs are not publicly available. 
This situation makes it very challenging for new researchers to further study this problem since designing and implementing a decent NLTH AI is often very complicated and tedious. 
To fill this gap, in OpenHoldem, we design and implement four different types of NLTH AIs, which are strong enough to serve as a good starting point for future research in this area.

\subsubsection{Rule Based AI}
The rule-based method is probably the most straightforward way to implement NLTH AI. 
A rule-based NLTH AI consists of a collection of rules designed by domain experts. 
In OpenHoldem, we develop $\mathcal{A}^\mathcal{R}$, a strong rule-based NLTH AI designed by some skilled Texas Hold'em players in our research group. 
Our rule-based AI $\mathcal{A}^\mathcal{R}$ handles about $10^6$ different scenarios that are likely to occur in the real play of NLTH and contains tens of thousands of lines of code. 
As a suggestion, when researchers implement their own NLTH AIs, it is useful to compare them to our rule-based AI $\mathcal{A}^\mathcal{R}$ as a sanity check.

\begin{table*}[htbp]
	\centering
	\caption{OpenHoldem provides many rule-based AIs with different styles and strengths.}
		\resizebox{1\linewidth}{!}{%
		\begin{tabular}{c|c|l}
			\hline
			\hline
			NLTH AI Name & Exploitability  & Description  \\ \hline
			CallAgent  & Very High	  &  Always Call/Check.  \\ \hline
			ManiacAgent  & Very High	  &  Always raise by half or one pot randomly.  \\ \hline
			RandomAgent  &  High   & Randomly select legal actions.  \\ \hline
			TimidAgent  &  High	  &  Calls when holding the nut; else folds to any bet.  \\ \hline
			CandidAgent  &  High	  &  Bets 1/4 to one pot depending on hand strength, checks/calls with marginal hands, folds weak hands.  \\ \hline
			FickleAgent  &  High	  &  Randomly change the strategy every $N$ hands.  \\ \hline
			LooseAggressiveAgent  &  High	  &  Bets/raises aggressively with a wide range of hands.  \\ \hline
			LoosePassiveAgent	  &  High	 & Calls with most hands, folds weak hands, rarely raises.
			  \\ \hline
			TightPassiveAgent  &  High	  &  Calls with good hands, folds most hands, rarely raises.
			  \\ \hline
			TightAggressiveAgent  &  Moderate	 &   Similar to CandidAgent, with refined hand ranges and bluffing.  \\ \hline
			$\mathcal{A}^\mathcal{R}$  & Low	  &  A relatively strong rule AI designed by using the knowledge of some skilled Texas Hold'em players.  \\ \hline
		\end{tabular}%
			}
	\label{rule_ais}
\end{table*}

Besides the strong rule-based AI $\mathcal{A}^\mathcal{R}$, we also designed some other rule-based AIs with different styles and strengths (Table~\ref{rule_ais}).
These agents can be used as learning materials for beginners, and more importantly, they can also help researchers to carry out research on opponent modeling. 
These rule-based AIs calculate the expected winning probability at each stage, and then make decisions based on these probabilities and different predefined rules.

\subsubsection{CFR Based Static AI}
While the rule-based approach provides a simple framework for implementing NLTH AIs, the resulting strategies are exploitable. 
Therefore, most recent studies in NLTH AIs are focused on approximating the theoretically unexploitable Nash equilibrium strategies. 
Among them, the most successful approach is the CFR algorithm~\cite{zinkevich2008regret} and its various variants~\cite{tammelin2014solving,jackson2016compact,brown2019solving}. 
CFR type algorithms iteratively minimizes the regrets of both players so that the time-averaged strategy gradually approximates the Nash equilibrium.
In OpenHoldem, we design and implement $\mathcal{A}^\mathcal{C}$, a strong CFR based NLTH AI, which aims to serve as a starting point for the large-scale equilibrium-finding research. 
Overall, $\mathcal{A}^\mathcal{C}$ first uses the abstraction algorithm to create a smaller abstract game, then approximates the Nash equilibrium strategy in this abstract game, and finally executes the resulting strategy in the original game.

The abstraction algorithm aims to take a large-scale imperfect information game as input and output a smaller but strategically similar game that is solvable by current equilibrium-finding algorithms. 
It usually consists of two parts, information abstraction and action abstraction. 
In $\mathcal{A}^\mathcal{C}$, we use the potential-aware information abstraction algorithm~\cite{ganzfried2014potential}, which uses the k-means algorithm with the earth mover's distance metric to cluster cards with similar potential. 
Action abstraction further reduces the size of the game tree by restricting the available actions, which is especially important in games with large action spaces, such as NLTH. 
In $\mathcal{A}^\mathcal{C}$, we restrict the actions to Fold, Call/Check, Bet Half Pot, Bet Pot, and All-In.

\begin{algorithm}[tb]
	\caption{The CFR+ algorithm which is used to train $\mathcal{A}^\mathcal{C}$.}
	\label{cfr}
	\textbf{Input}: The abstract game $\mathcal{G}$, the randomly initialized strategy profile ${\sigma}^{1}$, the zero initialized cumulative regret $R^0$ and cumulative strategy $S^0$.   \\
	\textbf{Parameter}: The number of iterations $T$.\\
	\textbf{Output}: The approximate Nash equilibrium $\bar{\sigma}^{T}=\{\bar{\sigma}^{T}_1,\bar{\sigma}^{T}_2\}$.
	\begin{algorithmic}[1] 
		\For{$t = 1 \to T$}
		\For{$i = 1 \to 2$}
		\State 
		$v^{\sigma^t}_{i}(h)=\sum_{h \sqsubseteq z, z \in \mathcal{Z}} \pi^{\sigma^t}_{-i}(h) \pi^{\sigma^t}(h, z)u_{i}(z)$
		\State
		$v^{\sigma^t}_{i}(a|h)=v^{\sigma^t}_{i}(ha)$
		\State 
		$v^{\sigma^t}_{i}(I_{i})=\sum_{h \in I_{i}}v^{\sigma^t}_{i}(h)$
		\State
		$v^{\sigma^t}_{i}(a|I_{i})=\sum_{h \in I_{i}}v^{\sigma^t}_{i}(ha)$
		\State
		$r^{\sigma^t}_{i}(a|I_{i})=v^{\sigma^t}_{i}(a|I_{i})-v^{\sigma^t}_{i}(I_{i})$
		\State
		$R^{t}_{i}(a|I_{i}) = \text{max}(0,R^{t-1}_{i}(a|I_{i}) + r^{\sigma^{t}}_{i}(a|I_{i}))$
		\State
		$\sigma^{t+1}_{i}(a|I_{i})=
		\nicefrac{R^{t}_{i}(a|I_{i})}{\sum_{a \in \mathcal{A}(I_{i})}R^{t}_{i}(a|I_{i})}$
		\State
		$ S^{t}_i(a|I_{i})=S^{t-1}_i(a|I_{i}) + \pi^{\sigma^{t}}_{i}(I_{i}) \sigma^{t}_{i}(a|I_{i})$
		\EndFor
		\EndFor
		\State
		$\bar{\sigma_{i}}^{T}(a|I_{i})=\nicefrac{S^{T}_i(a|I_{i})}{\sum_{a \in \mathcal{A}(I_{i})} S^{T}_i(a|I_{i})  }  $
	\end{algorithmic}
\end{algorithm}

After obtaining the manageable abstract game $\mathcal{G}$, we use the iterative CFR+~\cite{tammelin2014solving} algorithm to approximating the Nash equilibrium in $\mathcal{G}$. 
As shown in Algorithm~\ref{cfr}, given the current strategy profile $\sigma^t$, we first calculate the cumulative regret of each action after $t$ iterations in Line $8$. 
Then, the new strategy in the $t+1$-th iteration is updated in Line $9$ by the regret-matching algorithm. 
Finally, by normalizing the cumulative strategy $S^{T}$ in Line $13$, the average strategy $\bar{\sigma}^{T}$ will approach a Nash equilibrium when $T$ is large enough. 
During the actual play phase, $\mathcal{A}^\mathcal{C}$ first finds the abstract state that corresponds to the current real state of the game. 
Then, the approximate Nash equilibrium $\bar{\sigma}^{T}$ of the abstract game is queried for the probability distribution over different actions. 
Finally, an action is sampled from this distribution and played in the actual game, if applicable.

\subsubsection{DeepStack-Like Online AI}
In essence, the $\mathcal{A}^\mathcal{C}$ agent is a static table calculated offline that contains the probability distributions over possible actions in all situations. 
During actual play, if the opponent chooses an action that is not in the action abstraction of $\mathcal{A}^\mathcal{C}$, \ie, an off-tree action, $\mathcal{A}^\mathcal{C}$ round this off-tree action to a nearby in-abstraction action. 
A more principled approach to calculate the off-tree action's response is by solving a subgame that immediately follows that off-tree action. 
DeepStack~\cite{moravvcik2017deepstack} is a representative online algorithm based on this idea. 
In particular, DeepStack allows computation to be focused on specific situations raised when making decisions using a sound local strategy computation algorithm called continual re-solving. 
To make continual re-solving computationally tractable, DeepStack replaces sub-trees beyond a certain depth with a learned value function based on deep neural network.

The authors of DeepStack~\cite{moravvcik2017deepstack} does not release the training code or model for NLTH. 
They only release a pedagogical code for Leduc Hold'em\footnote{\url{https://github.com/lifrordi/DeepStack-Leduc}} which cannot be transferred directly to NLTH because the game tree of NLTH is much larger than that of Leduc Hold'em, and the pedagogical code does not contain the necessary acceleration techniques for NLTH. 
Based on this situation, we reimplement DeepStack for NLTH following the original paper's key ideas and obtain an online AI called $\mathcal{A}^\mathcal{D}$, which aims to serve as a starting point for the research of subgame solving in large-scale imperfect-information games. 
Specifically, we spend several weeks using $120$ GPUs to generate millions of training samples for the river, turn, and flop value networks. 
Each training sample is generated by running $1000$ CFR+ iterations based on a random reach probability. 
Since generating these training data requires huge computing resources, we will provide download links for these training data later. 
Everyone can freely use these data for research.
It is worth noting that Noam Brown, the creator of Libratus, recently co-authored a paper~\cite{zarick2020unlocking}, in which they also reimplemented DeepStack.
$\mathcal{A}^\mathcal{D}$ has achieved similar results to theirs, which validates the correctness of our reimplementation.

\subsubsection{Deep Reinforcement Learning Based AI}
The three agents, \ie, the rule-based AI $\mathcal{A}^\mathcal{R}$, the CFR based static AI $\mathcal{A}^\mathcal{C}$, and the DeepStack-like online AI $\mathcal{A}^\mathcal{D}$, described in the previous sections are all based on improvements of existing techniques.
These AIs often rely on different kinds of NLTH domain knowledge, such as expert rules in $\mathcal{A}^\mathcal{R}$ and handcrafted abstraction algorithms in $\mathcal{A}^\mathcal{C}$.
Besides, there are also computational issues, \ie, in the inference stage of $\mathcal{A}^\mathcal{D}$, the CFR iteration process consumes much computation.
Specifically, to ensure $\mathcal{A}^\mathcal{D}$'s high-quality prediction, this iteration process often needs to be carried out for more than 1,000 times in practice.

Based on the above considerations, in OpenHoldem, we further propose a high-performance and lightweight NLTH AI, \ie, $\mathcal{A}^\mathcal{RL}$, obtained with an end-to-end deep reinforcement learning framework.
$\mathcal{A}^\mathcal{RL}$ adopts a pseudo-Siamese architecture to directly learn from the input state information to the output actions by competing the learned model with its different historical versions. 
The main technical contributions of $\mathcal{A}^\mathcal{RL}$ include a novel state representation of card and betting information, a novel reinforcement learning loss function, and a new self-play procedure to generate the final model.
We finish the training of $\mathcal{A}^\mathcal{RL}$ in three days using only one single computing server of 8 GPUs and 64 CPU cores. 
During inference, $\mathcal{A}^\mathcal{RL}$ takes only \(3.8\times10^{-3}\) second for each decision in a single-core CPU of 2.00GHz.
$\mathcal{A}^\mathcal{RL}$ is the first AI that obtains competitive performance in NLTH solely through RL.

\paragraph{The Overall Architecture}
$\mathcal{A}^\mathcal{RL}$ aims to remove the expensive computation of CFR iteration in both the training and testing stages of a NLTH AI while eliminating the need of domain knowledge. 
It thus pursues an end-to-end learning framework to perform efficient and effective decision-making in imperfect-information games. 
Here \emph{end-to-end} means that the framework directly accepts the game board information and outputs the actions without encoding handcrafted features as inputs or performing iterative reasoning in the decision process. 
$\mathcal{A}^\mathcal{RL}$ adopts the RL framework to achieve this goal, and the only force to drive the model to learn is the reward.

\begin{figure}[t]
	\centering
	{
		{\includegraphics[width=1.0\linewidth]{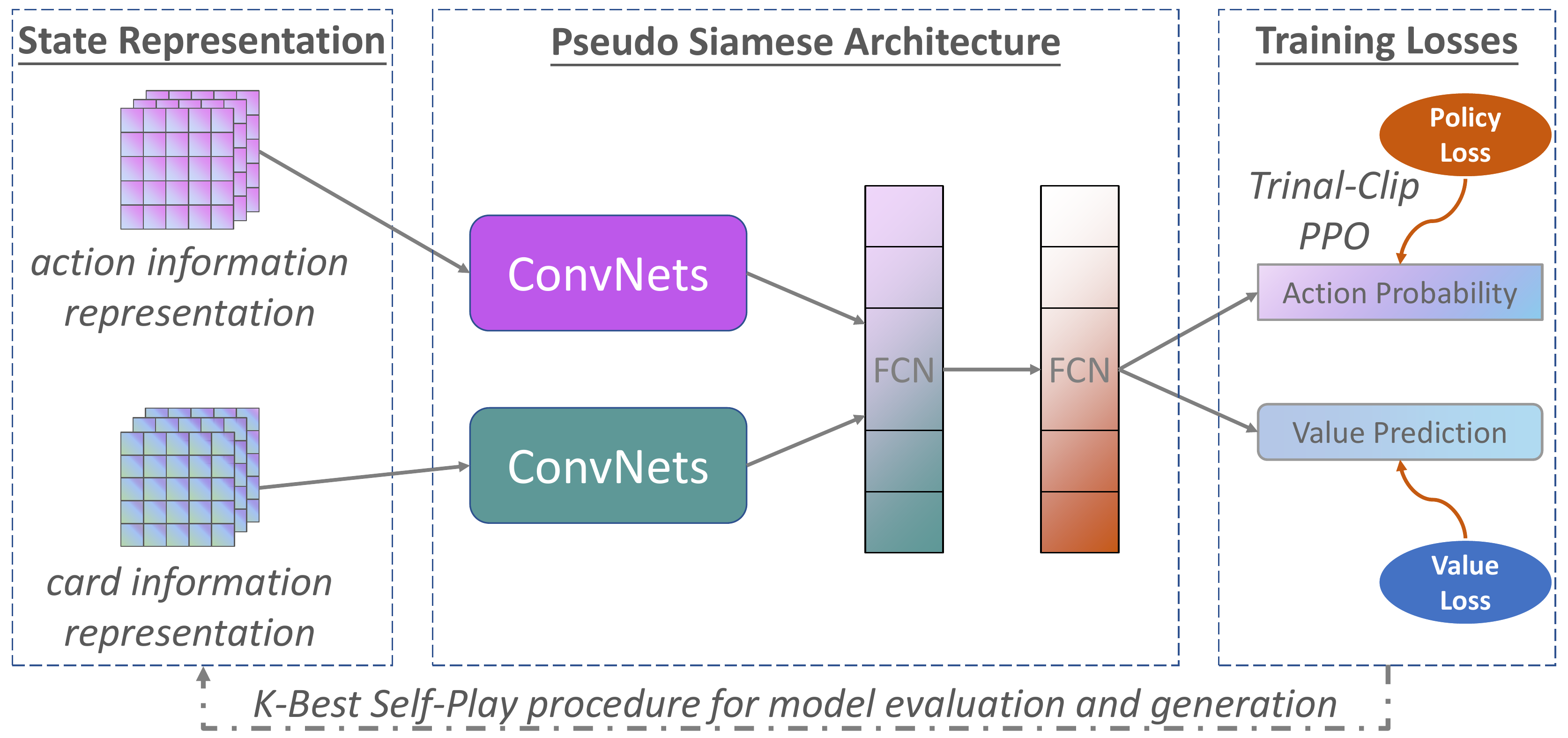}}
		\caption{End-to-end learning architecture of our deep RL based AI $\mathcal{A}^\mathcal{RL}$.}
		\label{overall_archi}
	}
\end{figure}

In NLTH, the game board information includes the current and historical card information and the player action information. 
The agent chooses from a set of betting actions to play the game and try to win more rewards. 
To capture the complex relationship among the game board information, the desired betting actions, and the game rewards, we design a pseudo-Siamese architecture equipped with the RL schema to learn the underlying relationships from end to end. 
We illustrate the end-to-end learning architecture of $\mathcal{A}^\mathcal{RL}$ in Figure~\ref{overall_archi}.

As shown in Figure~\ref{overall_archi}, the input of the architecture is the game state representations of action and card information, which are respectively sent to the top and bottom streams of the Siamese architecture. 
Since the action and card representations provide different kinds of information to the learning architecture, we first isolate the parameter-sharing of the Siamese architecture to enable the two ConvNets to learn adaptive feature representations, which are then fused through fully connected layers to produce the desired actions. 
This design is the reason why we call it pseudo-Siamese architecture. 
To train $\mathcal{A}^\mathcal{RL}$, we present a novel Trinal-Clip loss function to update the model parameters using RL algorithms. 
We obtain the final model through a new self-play procedure that plays the current model with a pool of its \(K\) best historical versions to sample diverse training data from the huge game state space.
We believe these new techniques and underlying principles are helpful to develop general learning algorithms for more imperfect-information games.

\paragraph{Effective Game State Representation}
The existence of private information and flexibility of bet size cause the NLTH AI learning extremely challenging. 
To obtain an effective and suitable feature representation for end-to-end learning from the game state directly to the desired action, we design a new multi-dimensional feature representation to encode both the current and historical card and bet information.

\begin{figure}[t]
	\centering
	\includegraphics[width=1.0\linewidth]{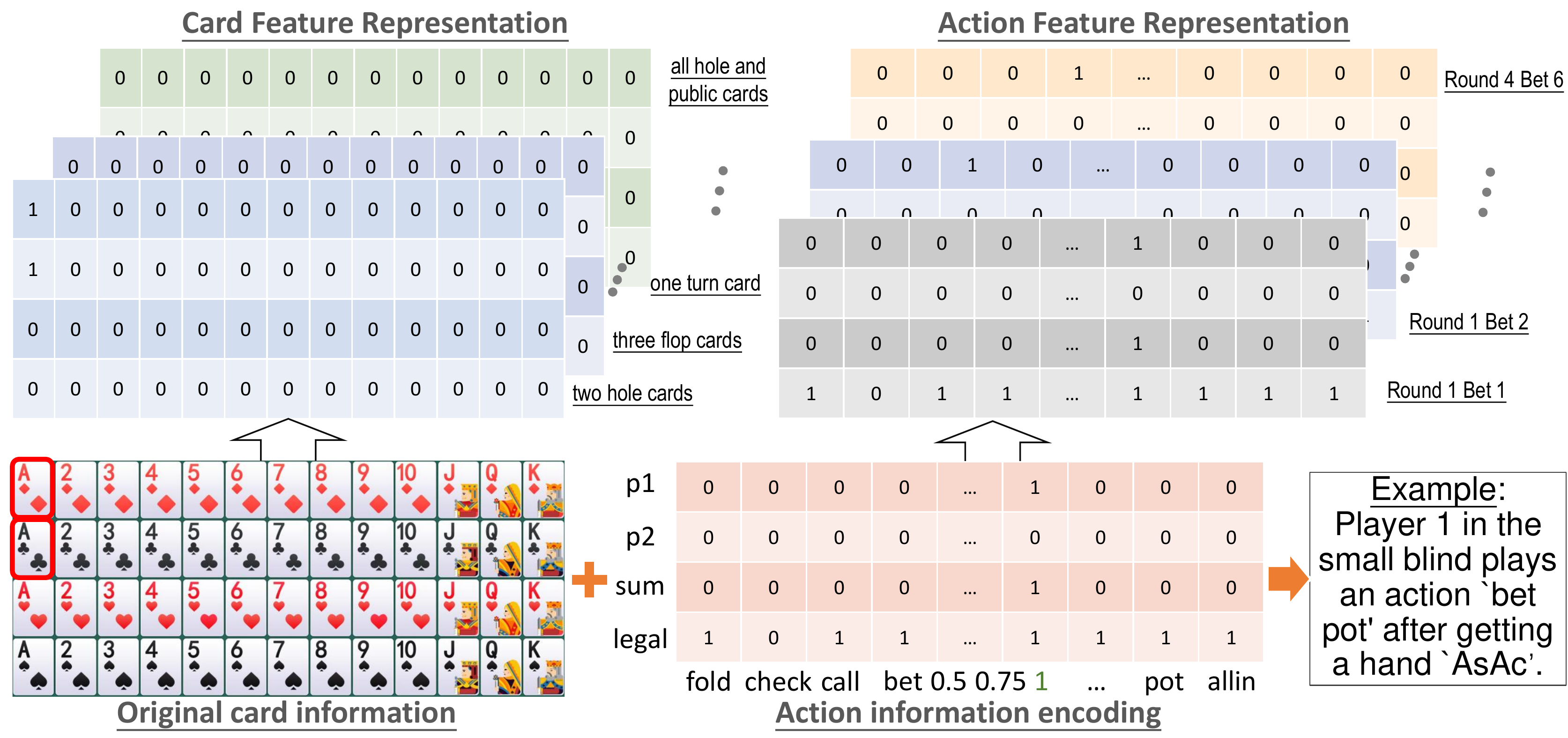}
	\caption{A state representation example when Player 1 in the small blind plays `bet pot' after getting an hand `AsAc'.}
	\label{state_representation}
\end{figure}

In NLTH, the card and action information exhibit different characteristics. 
We thus represent them as two separated three-dimension tensors and let the network learn to fuse them (Figure~\ref{overall_archi}). 
We design the card tensor in six channels to represent the agent's two private cards, three flop cards, one turn card, one river card, all public cards, and all private and public cards. 
Each channel is a \(4\times13\) sparse binary matrix, with 1 in each position denoting the corresponding card. 
For the action tensor, since there are usually at most six sequential actions in each of the four rounds, we design it in 24 channels. 
Each channel is a \(4\times n_b\) sparse binary matrix, where \(n_b\) is the number of betting options, and the four dimensions correspond to the first player's action, the second player's action, the sum of two player's action, and the legal actions. 
To understand this representation, Figure~\ref{state_representation} illustrates one example that a player in the small blind plays an action `bet pot' after getting a hand `AsAc'.

This representation has several advantages: 1) there is no abstraction of the card information thus reserves all the game information; 
2) the action representation is general and can denote different number of betting options (though \(n_b=9\) produce satisfactory results in the experiment); 
3) all the historical information is encoded to aid reasoning with hidden information; 
and 4) the multi-dimensional tensor representation is very suitable for modern deep neural architectures like ResNet~\cite{CVPR16ResNet} to learn effective feature hierarchies, as verified in the AlphaGo AI training.

\paragraph{Effective Learning with Trinal-Clip PPO}
With the multi-dimensional feature representation, a natural choice is to use the current state-of-the-art reinforcement learning algorithms such as PPO~\cite{schulman2017proximal} to train the deep architecture.
PPO is an actor-critic framework which trains a value function \(V_{\theta}(s_t)\) and a policy \(\pi_{\theta}(a_t|s_t)\).
PPO defines a ratio function $r_t(\theta) = \tfrac{\pi_{\theta}(a_t|s_t)}{\pi_{\theta'}(a_t|s_t)}$ as the ratio between the current policy $\pi_{\theta}$ and the old policy $\pi_{\theta'}$, and a policy loss function $\mathcal{L}^{p}$ as:
\begin{equation}
	\small{
		\mathcal{L}^{p}(\theta)=\mathbb{E}_{t}\left[\min \left(r_{t}(\theta) \hat{A}_{t}, \operatorname{clip}\left(r_{t}(\theta), 1-\epsilon, 1+\epsilon\right) \hat{A}_{t}\right)\right],
	}
\end{equation}
where $\hat{A_t}$ is the advantage function, \(\operatorname{clip}(r_{t}(\theta), 1-\epsilon, 1+\epsilon)\) ensures \(r_t\) lie in the interval \((1-\epsilon, 1+\epsilon)\), and \(\epsilon\) is a clip ratio hyper-parameter with typical value 0.2.
PPO's value loss $\mathcal{L}^{v}$ is defined as:
\begin{equation}
	\mathcal{L}^{v}(\theta)=\mathbb{E}_{t}\left[\left(R_{t}^{\gamma}-{V}_{\theta}(s_t)\right)^{2}\right],
\end{equation}
in which $R_{t}^{\gamma}$ represents the traditional $\gamma$-return~\cite{sutton2018reinforcement}.

However, the above PPO loss function is difficult to converge for NLTH AI training.
We find two main reasons for this problem: 1) when $\pi_{\theta}(a_t|s_t)\gg\pi_{\theta^{'}}(a_t|s_t)$ and the advantage function $\hat{A_t} \textless 0$, the policy loss $\mathcal{L}^{p}(\theta)$ will introduce a large variance; 2) due to the strong randomness of NLTH, the value loss $\mathcal{L}^{v}(\theta)$ is often too large.
To speed up and stabilize the training process, we design a Trinal-Clip PPO loss function. 
It introduces one more clipping hyper-parameter $\delta_1$ for the policy loss when $\hat{A_t} \textless 0$, and two more clipping hyper-parameters $\delta_2$ and $\delta_3$ for the value loss. 
The policy loss function $\mathcal{L}^{tcp}$ for Trinal-Clip PPO is defined as:
\begin{equation}
	\mathcal{L}^{tcp}(\theta)\!\!=\!\!\mathbb{E}_{t}\!\!\left[\!\operatorname{clip}\left(r_{t}(\theta), \operatorname{clip}\left(r_{t}(\theta), 1\!-\!\epsilon, 1\!+\!\epsilon\right)\!,\delta_1\!\right) \hat{A}_{t}\!\right],
\end{equation}
where $\delta_1 > 1+\epsilon$, and $\epsilon$ is the original clip in PPO.
The clipped value loss function $\mathcal{L}^{tcv}$ for Trinal-Clip PPO is defined as:
\begin{equation}
	\mathcal{L}^{tcv}(\theta)=\mathbb{E}_{t}\left[\left( \operatorname{clip}\left(R_{t}^{\gamma}, -\delta_2, \delta_3\right)-{V}_{\theta}(s_t)\right)^{2}\right],
\end{equation}
where \(\delta_2\) and \(\delta_3\) do not require manual tuning but represent the total number of chips the player and the opponent has placed, respectively. 
\(-\delta_2\) represent the state value when the player folds, similarly, \(\delta_3\) is the state value when the opponent folds.
This value-clip loss significantly reduces the variance during the training process.
Our proposed Trinal-Clip PPO loss function improves the learning effectiveness of the actor-critic framework, and we believe it is applicable for a wide range of RL applications with imperfect information.

\paragraph{Efficient Self-Play Procedure}
With the proposed Trinal-Clip PPO loss function, the most direct way is using the self-play algorithm~\cite{Samuel59SelfPlay} to train the NLTH agent.
However, due to the private information in NLTH, simple self-play learning designed for perfect information games~\cite{silver2016mastering,silver2018general} often causes the agent trapped in a local minimum and defeated by agents with counter-strategies.
AlphaStar~\cite{vinyals2019grandmaster} designs a population-based training (PBT) procedure to maintain multiple self-play agents and obtains excellent results in the real-time strategy game StarCraft \uppercase\expandafter{\romannumeral2}. 
However, the PBT procedure needs a tremendous computational resource to ensure good performance.

To obtain a high-performance NLTH AI with both low computation cost and strong decision-making ability, we propose a new type of self-play algorithm which trains only one agent but learns strong and diverse policies.
The proposed algorithm maintains a pool of competing agents from the historical versions of the main agent.
Then, by competing among different agents, the algorithm selects the \(K\) best survivors from their ELO~\cite{vinyals2019grandmaster} scores and generates training data simultaneously.
The main agent learns from the data and thus can compete with different opponents, maintaining a strong decision-making ability of high-flexible policies.
Since the proposed algorithm performs self-play among the main agent and its \(K\) best historical versions, we refer to it as \(K\)-Best Self-Play.
Our proposed \(K\)-Best Self-Play inherits PBT's merit of diverse policy styles while maintains computational efficiency of single-thread agent training, striking a good balance between efficiency and effectiveness.

\subsection{Online Testing Platform}
\begin{figure}[htbp]
	\centering
	\includegraphics[width=1\linewidth]{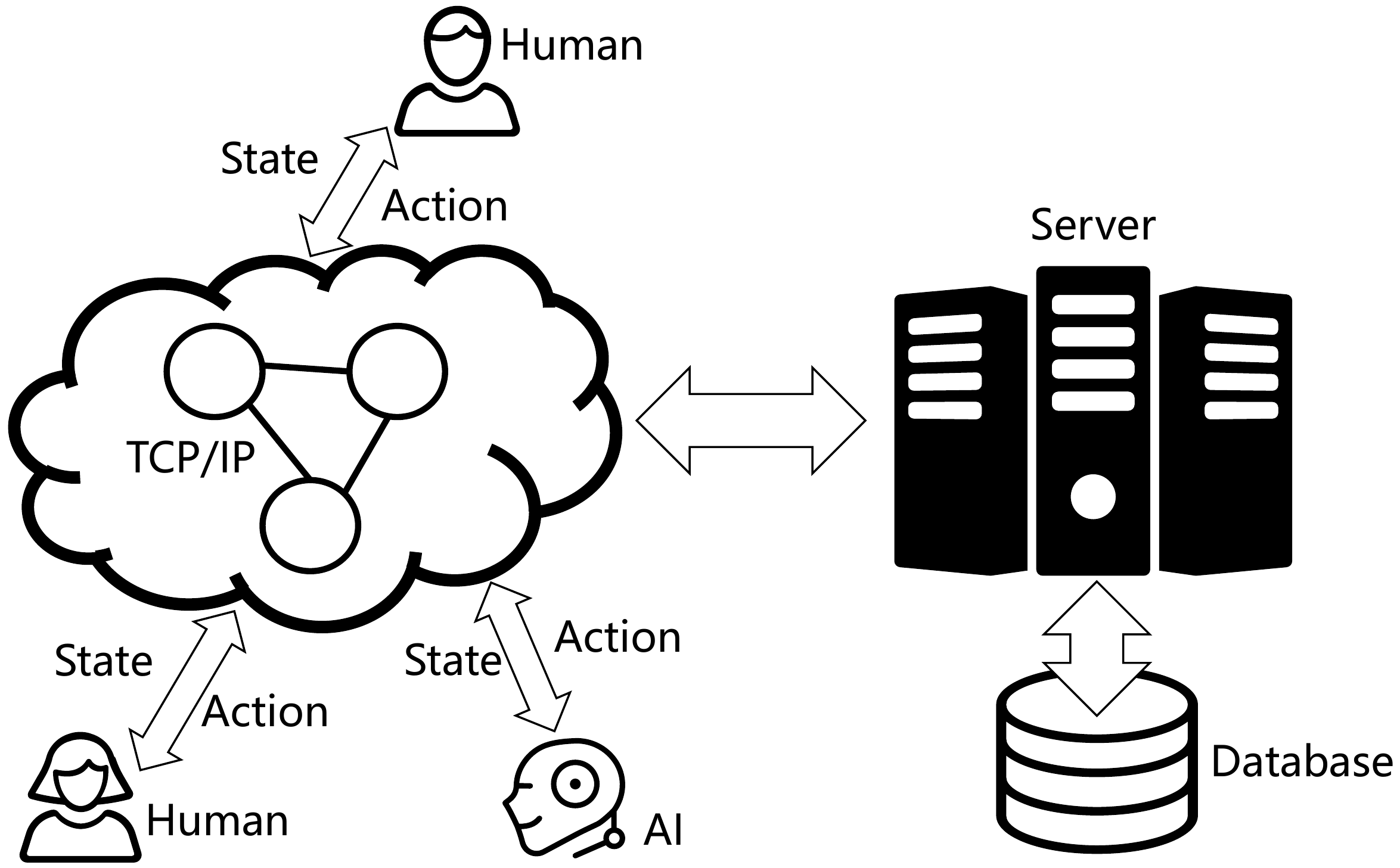}
	\caption{The schematic diagram of our testing platform's system architecture.}
	\label{platform}
\end{figure}

In order to make the comparisons between different NLTH AIs easier, we develop an online testing platform with the above four strong baseline AIs, \ie, $\mathcal{A}^\mathcal{R}$, $\mathcal{A}^\mathcal{C}$, $\mathcal{A}^\mathcal{D}$ and $\mathcal{A}^\mathcal{RL}$ built-in.
Researchers can compare the performances between their own AIs and the built-in baselines through easy-to-use APIs. 
Figure~\ref{python_code} shows an example Python code of connecting to the platform for testing NLTH AIs. 
The NLTH AI designers only need to implement one function, \ie, \texttt{act}, without caring about the internal structure of the platform. 
The input of \texttt{act} is the current game state, which is obtained from the platform through TCP sockets. 
The output of \texttt{act} is the action to take in the current game state according to the designer's algorithm. 
The output action is also sent to the platform through TCP sockets. 
Figure~\ref{platform} shows the system architecture of our testing platform.
The server is responsible for playing the poker hands according to the rules of NLTH. 
It also dynamically schedules requests and allocates resources when necessary. 
Our platform not only supports testing between different AIs, but also between humans and AIs.

\begin{figure}[t]
	\lstset{language=Python}
	\lstset{frame=lines}
	\lstset{breaklines=true}
	\lstset{label={lst:code_direct}}
	\lstset{basicstyle=\ttfamily\small}
	\begin{lstlisting}
import json
import socket
...

# The IP address and port of the platform
server_ip = '127.0.0.1'
server_port = 1080

# Create socket and connect to the platform
client = socket.socket(socket.AF_INET, socket.SOCK_STREAM)
client.connect(server_ip, server_port)

while True:
  # Get state in json format from the platform
  state = recvJson(client)
  ...
  # Use your awesome AI to get the action
  action = act(state)
  ...
  # send your action to the platform
  sendJson(client, action)

# Close the socket
client.close()
	\end{lstlisting}
	\caption{An example Python code of connecting to the platform for testing NLTH AIs.}
	\label{python_code}
\end{figure}

We are more than happy to accept high-performance AIs submitted by everyone to continuously enrich the baseline AIs of OpenHoldem, with the ultimate goal of providing an \emph{\textbf{NLTH AI Zoo}} for the research community.
Currently, there are dozens of NLTH AI researchers and developers are using this platform. 
It has accumulated about 20 million high-quality poker data and the data increases by about 100,000 per day. 
We believe that these large-scale data will also facilitate the research of data-driven imperfect-information game solving, imitation learning and opponent modeling algorithms.

\section{Experiments}
In this section, we first compare the performance of our baseline NLTH AIs with other publicly available NLTH AIs using the proposed evaluation protocols and online testing platform. 
Then, we conduct a set of ablation studies to analyze the effects of various design choices in the baseline NLTH AIs.

\subsection{Comparison to the State-of-the-Arts}
To the best of our knowledge, Slumbot~\cite{jackson2013slumbot}, the champion of the 2018 Annual Computer Poker Competition (ACPC), is the only publicly available NLTH AI that provides comparisons through an online website\footnote{\url{https://www.slumbot.com/}}. 
Slumbot is a strong CFR-based agent whose entire policy is precomputed and used as a lookup table.
Similar to our $\mathcal{A}^\mathcal{C}$, Slumbot first uses some abstraction algorithm to create a smaller abstract NLTH game. 
Then it approximates the Nash equilibrium in the abstract game using the CFR-type algorithm and finally executes the resulting strategy in the original game.

\begin{table}[t]
	\centering
	\caption{The head-to-head performances (mbb/h) of the rule based AI $\mathcal{A}^\mathcal{R}$, the CFR based AI $\mathcal{A}^\mathcal{C}$, the DeepStack-like AI $\mathcal{A}^\mathcal{D}$, and the reinforcement learning based AI  $\mathcal{A}^\mathcal{RL}$ when playing against Slumbot, respectively.}
	\begin{tabular}{c|c|c|c|c}
		\hline
		\hline
		Baseline NLTH AIs & $\mathcal{A}^\mathcal{R}$ & $\mathcal{A}^\mathcal{C}$ & $\mathcal{A}^\mathcal{D}$ & $\mathcal{A}^\mathcal{RL}$ \\ \hline
		Performance (mbb/h) & 57 & -20 & 103  & 111 \\ \hline
	\end{tabular}
	\label{ResultWithSlumbot}
\end{table}

The original intention of Slumbot's website is to facilitate human players to compete with it, and there are no open source tools available to test the performance of AI against Slumbot. 
Due to the poor stability of Slumbot's website, the way of playing with a simulated browser will lose the connection after a certain number of matches, so we develop a software which use an alternative method of sending data packets directly. 
Based on this software\footnote{We will open source this tool in OpenHoldem.}, we compare each of our baseline NLTH AIs with Slumbot for 100,000 hands, and the head-to-head based evaluation results (AIVAT) are shown in Table~\ref{ResultWithSlumbot}. 

We can see that both the DeepStack-like AI $\mathcal{A}^\mathcal{D}$ and the reinforcement learning based AI $\mathcal{A}^\mathcal{RL}$ outperform Slumbot by a large margin. 
Although the performance of the CFR based AI $\mathcal{A}^\mathcal{C}$ is not as good as that of Slumbot, its performance is also commendable because Slumbot exploits a far more fine-grained abstraction algorithm.
An interesting result is that the rule-based AI $\mathcal{A}^\mathcal{R}$ outperforms Slumbot.
This result is not surprising, as it has been reported that the abstraction-based programs from the Annual Computer Poker Competition are exploitable~\cite{lisy2017eqilibrium}.
These experimental results illustrate that our baseline NLTH AIs are adequate to serving as a good starting point for NLTH AI research.

The DeepStack-like AI $\mathcal{A}^\mathcal{D}$ and the RL based AI $\mathcal{A}^\mathcal{RL}$ obtain the best performance among the four baselines. 
They are also the most complicated baselines in terms of design and implementation. 
Next, We conduct some ablation studies to understand the effects of their various design choices.

\subsection{Ablation Study on $\mathcal{A}^\mathcal{D}$}

\subsubsection{The Effects of Training Data Size}

\begin{figure}[t]
	\centering
	\includegraphics[width=0.8\linewidth]{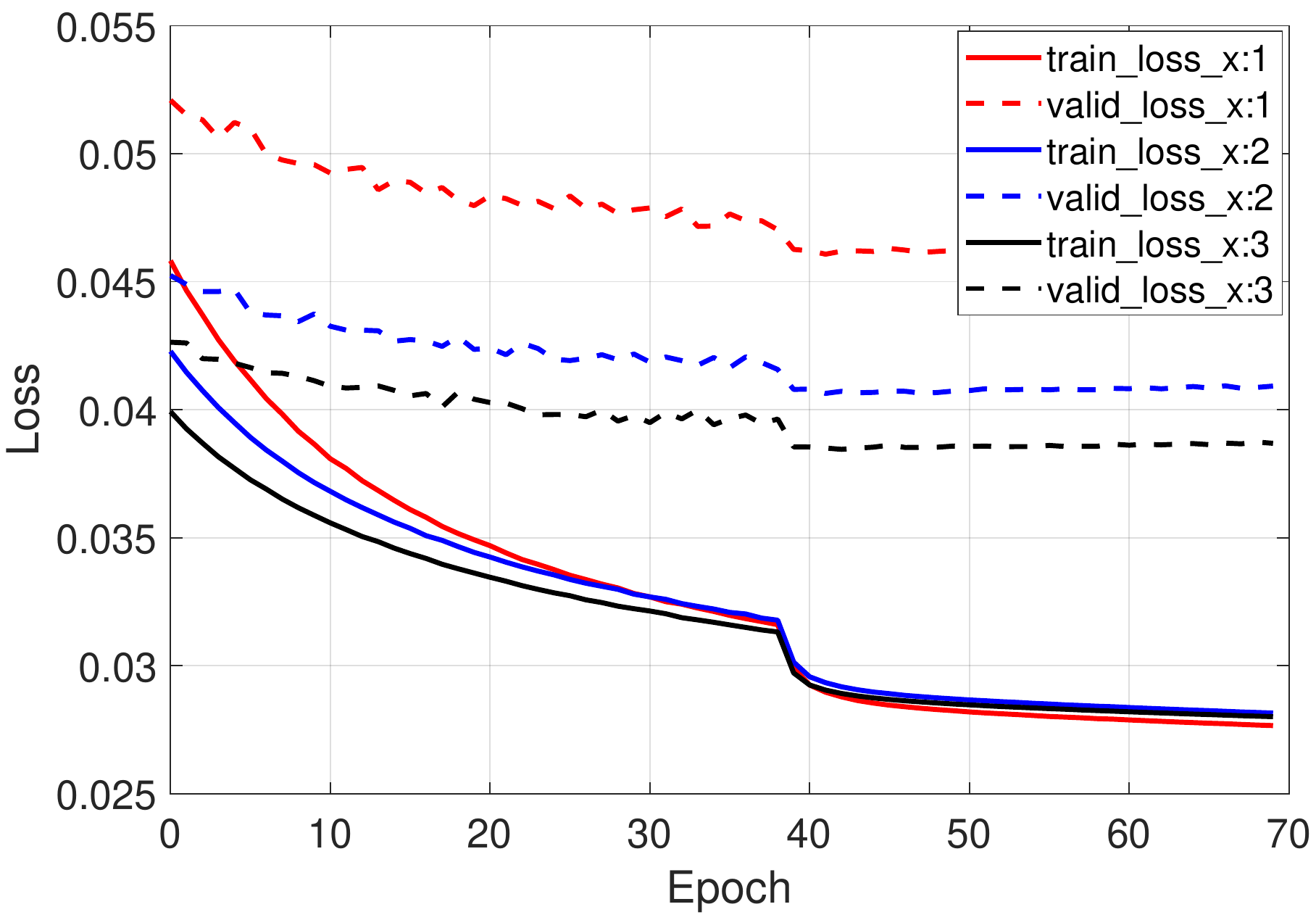}
	\caption{The training and validation loss curves of the flop network when using $x \in \{1,2,3\}$ million training samples, respectively.}
	\label{Loss}
\end{figure}

\begin{figure}[htbp]
	\centering
	\includegraphics[width=0.8\linewidth]{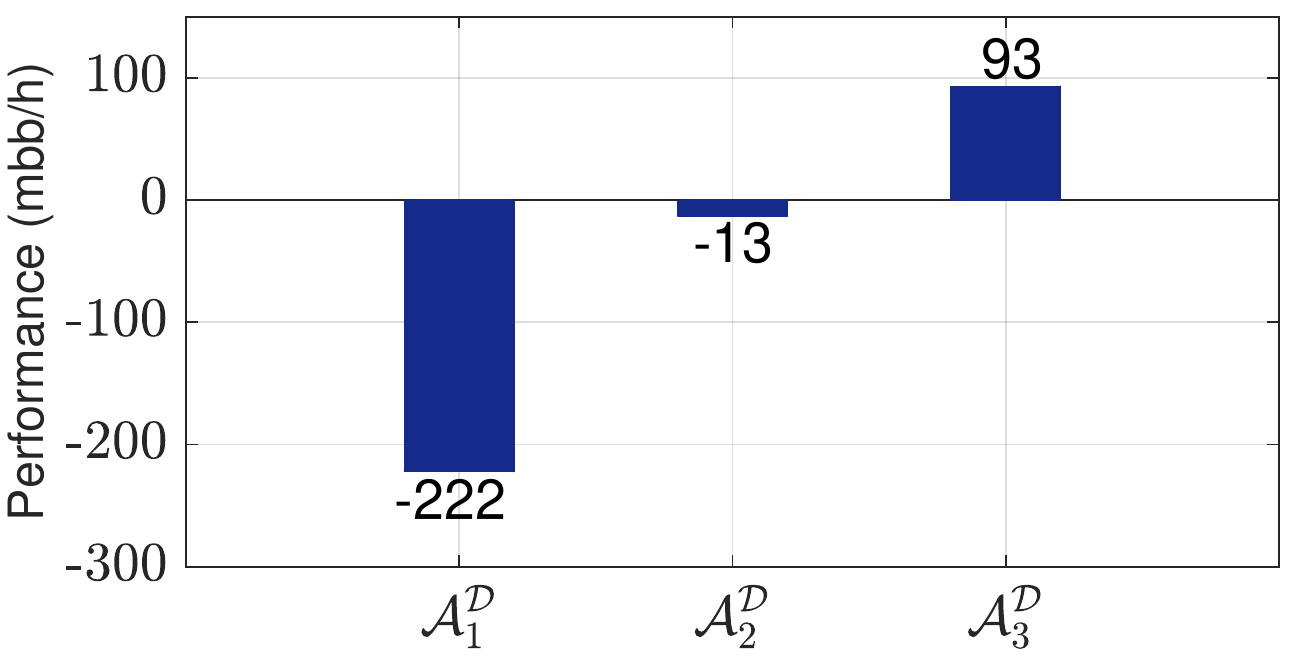}
	\caption{The head-to-head performances of
		$\mathcal{A}^\mathcal{D}_{1}$,
		$\mathcal{A}^\mathcal{D}_{2}$ and
		$\mathcal{A}^\mathcal{D}_{3}$ when playing against Slumbot, respectively.}
	\label{DeepStackWithSlumbot}
\end{figure}

The training of the river, turn, and flop value networks of $\mathcal{A}^\mathcal{D}$ requires a lot of training data. 
We use $\mathcal{A}^\mathcal{D}_{x}$ to denote the DeepStack-like NLTH AIs whose flop networks are obtained by training with $x$ million samples. 
Figure~\ref{Loss} shows the loss curves of the flop network during training when $x \in \{1,2,3\}$. 
It is clear that the flop network suffers from severe over-fitting when the training data size is small, and increasing the training data size alleviates this phenomenon. 
The head-to-head based evaluation results (AIVAT) in Figure~\ref{DeepStackWithSlumbot} also show that DeepStack-type AI is data-hungry and more training data results in a stronger AI.

\subsubsection{The Effects of CFR Iterations During Continual Re-solving} 
We use $\mathcal{A}^{\mathcal{D}:y}_{3}$ to denote the DeepStack-like NLTH AIs, which use $y$ CFR iterations during the continual re-solving procedure. 
We find that $\mathcal{A}^{\mathcal{D}:500}_{3}$ loses $224$ mbb to Slumbot per hand, while $\mathcal{A}^{\mathcal{D}:1000}_{3}$ wins Slumbot $93$ mbb per hand. 
These experimental results demonstrate that the number of CFR iterations during continual re-solving is critical to the performance of DeepStack-type AI.

\begin{figure*}[t]
	\centering
	\includegraphics[width=0.8\linewidth]{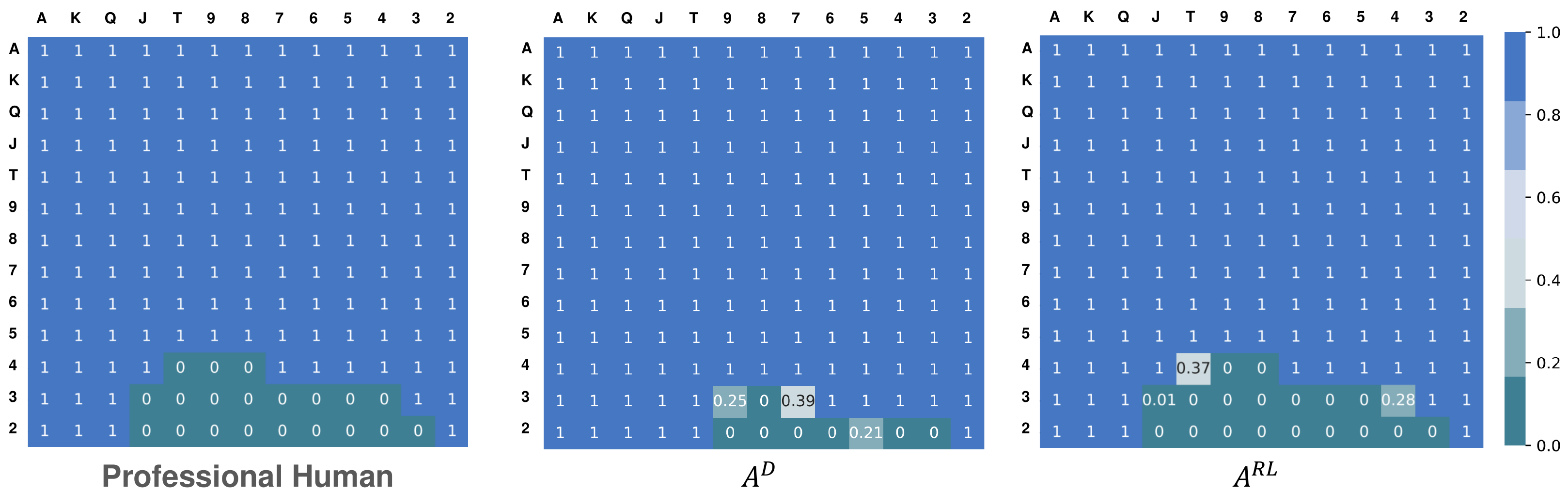}
	\caption{Probabilities for not folding as the first action for each possible hand. 
		The bottom-left half shows the policy when the suits of two private cards do not match, and the top-right half shows the policy when the suits of two private cards match. 
		Left to right represent the policies of Professional Human, $\mathcal{A}^\mathcal{D}$, and $\mathcal{A}^\mathcal{RL}$, respectively.}
	\label{policy_vis}
\end{figure*}

\subsection{Ablation Study on $\mathcal{A}^\mathcal{RL}$}

\begin{table}[t]
    \centering
	\caption{Ablation analyses of each component of $\mathcal{A}^\mathcal{RL}$.}
	\label{RLAgentAblation}
	\begin{tabular}{cccc}
		\hline
		\hline
		Name  &Training time (Hours) & ELO &   \\
		\hline
		Vector & $3.8$ & $78$  \\
		PokerCNN & $5.4$ & $359$  \\
		W/O History Information  & $6.3$ & $896$  \\
		\hline
		Original PPO Loss & $8.4$ & $1257$   \\
		Dual-Clip PPO Loss  & $8.4$ & $1308$   \\
		\hline
		Naive Self-Play  & $8.4$ & $1033$   \\
		Best-Win Self-Play  & $8.4$ & $1024$  \\
		Delta-Uniform Self-Play   & $8.6$ & $931$  \\
		PBT Self-Play  & $8.9$  & $892$  \\
		\hline
		$\mathcal{A}^\mathcal{RL}$  & $8.4$ & \textbf{1597}  \\
		\hline
	\end{tabular}
\end{table}

To analyze the effectiveness of each component of the RL based AI $\mathcal{A}^\mathcal{RL}$, we have conducted extensive ablation studies, as shown in Table~\ref{RLAgentAblation}.
The results of each row are obtained by replacing one component of $\mathcal{A}^\mathcal{RL}$, and the rest remains unchanged.
All models use the same number of training samples, and we use ELO scores to compare their performance.

\subsubsection{The Effects of Different State Representations}
For state representation comparison, we consider three alternative methods: 1) Vectorized state representation like DeepCFR~\cite{ICML19DeepCFR} (\emph{Vector}). 
It uses vectors to represent the card information (52-dimensional vectors) and the action information (each betting position represented by a binary value specifying whether a bet has occurred and a float value specifying the bet size);
2) PokerCNN-based state representation~\cite{yakovenko2016poker} (\emph{PokerCNN}) uses 3D tensors to represent card and action information together and use a single ConvNet to learn features;
3) State representation without history information (\emph{W/O History Information}) is similar to $\mathcal{A}^\mathcal{RL}$ except that it does not contain history action information.

As shown in Table~\ref{RLAgentAblation}, state representation has a significant impact on the final performance.
PokerCNN performs better than the vectorized state representation Vector, demonstrating that it is more effective to represent state information using structured tensors.
$\mathcal{A}^\mathcal{RL}$ outperforms PokerCNN since it uses a pseudo-Siamese architecture to handle card and action information separately.
$\mathcal{A}^\mathcal{RL}$ is also better than W/O History Information since historical action information is critical to decision-making in NLTH.
$\mathcal{A}^\mathcal{RL}$ obtains the best performance thanks to its effective multi-dimensional state representation, which encodes historical information and is suitable for ConvNets to learn effective feature hierarchies.

\subsubsection{The Effects of Different Loss Functions}
For the loss function, we evaluate $\mathcal{A}^\mathcal{RL}$'s Trinal-Clip PPO loss  against two kinds of PPO losses: 1) the Original PPO loss~\cite{schulman2017proximal} (\emph{Original PPO}); 2) the Dual-Clip PPO loss~\cite{ye2020mastering} (\emph{Dual-Clip PPO}).
As shown in Table~\ref{RLAgentAblation}, compared with the Original PPO, Dual-Clip PPO has a slight performance boost, and Trinal-Clip PPO ($\mathcal{A}^\mathcal{RL}$) obtains the best performance. 
This performance improvement is mainly because $\mathcal{A}^\mathcal{RL}$'s policy-clip and value-clip loss effectively limit its output to a reasonable range, thus ensuring the stability of the policy update.
In addition, we find the model with a small overall loss generally performs better after adding the value-clip loss, which is very convenient for model selection during training.

\subsubsection{The Effects of Different Self-Play Methods}
For self-play methods, we compare $\mathcal{A}^\mathcal{RL}$'s \emph{\(K\)-Best Self-Play} with 1) \emph{Naive Self-Play}~\cite{Samuel59SelfPlay}, which plays with the agent itself;
2) \emph{Best-Win Self-Play}~\cite{silver2016mastering}, which plays with the best agent in history;
3) \emph{Delta-Uniform Self-Play}~\cite{ICLR18DeltaSelfPlay}, which plays with the agent in the last \(\delta\) timestamps;
and 4) \emph{PBT Self-Play}~\cite{vinyals2019grandmaster}, which trains multiple agents and play with each other.
Interestingly, compared with the more sophisticated Delta-Uniform Self-Play and PBT Self-Play, Naive Self-Play and Best-Win Self-Play achieve better performance, possible because more complex self-play strategies are more data-hungry.
However, the performance of Naive and Best-Win Self-Play are still behind K-Best Self-Play, since simplistic self-play methods can not overcome the notorious cyclical strategy problem in imperfect-information games.
Our \(K\)-Best Self-Play method obtains the best performance under the same amount of training data, striking a good balance between efficiency and effectiveness.

\subsubsection{Exploitability Analysis}
We evaluate the exploitability of $\mathcal{A}^\mathcal{RL}$ with LBR.
However, we find that LBR fails to exploit $\mathcal{A}^\mathcal{RL}$, \ie, LBR loses to $\mathcal{A}^\mathcal{RL}$ by over 335.82 mbb/h in 40,000 hands.
While this result does not prove that $\mathcal{A}^\mathcal{RL}$ is flawless, it does demonstrate that $\mathcal{A}^\mathcal{RL}$ seeks to compute and play a low-exploitability strategy.
$\mathcal{A}^\mathcal{RL}$'s low exploitability is mainly attributed to its effective state representation, which encodes historical information to alleviate the partial observable problem and its efficient self-play strategy to address the game-theoretic challenges (\ie, cyclical strategy behavior) in imperfect-information games.

\subsubsection{Visualization of the Learned Policy}
To analyze $\mathcal{A}^\mathcal{RL}$'s learned policy, we compare the action frequencies where the agent is the first player to act and has no prior state influencing it~\cite{zarick2020unlocking} with those from human professional\footnote{Obtained from \url{https://www.wsop.com/how-to-play-poker/}} and $\mathcal{A}^\mathcal{D}$.
Figure~\ref{policy_vis} shows the policies on how to play the first two cards from the professional human and the two agents.
The polices of $\mathcal{A}^\mathcal{D}$ and $\mathcal{A}^\mathcal{RL}$ are very similar to those of the human professional, which further explains their good performance.

\section{Conclusion}
In this work, we present OpenHoldem, a benchmark for large-scale imperfect-information game research using NLTH. OpenHoldem provides an integrated toolkit with three main components: the comprehensive evaluation protocols, the strong baseline NLTH AIs, and an easy-to-use online testing platform. We plan to add more NLTH AIs to OpenHoldem in the future, with the ultimate goal of providing an NLTH AI Zoo for the research community. We hope OpenHoldem will facilitate further studies on the unsolved theoretical and computational issues in large-scale imperfect-information games.

\ifCLASSOPTIONcaptionsoff
  \newpage
\fi

\bibliographystyle{IEEEtran}
\bibliography{reference}

\begin{IEEEbiography}[{\includegraphics[width=1in,height=1.25in,clip,keepaspectratio]{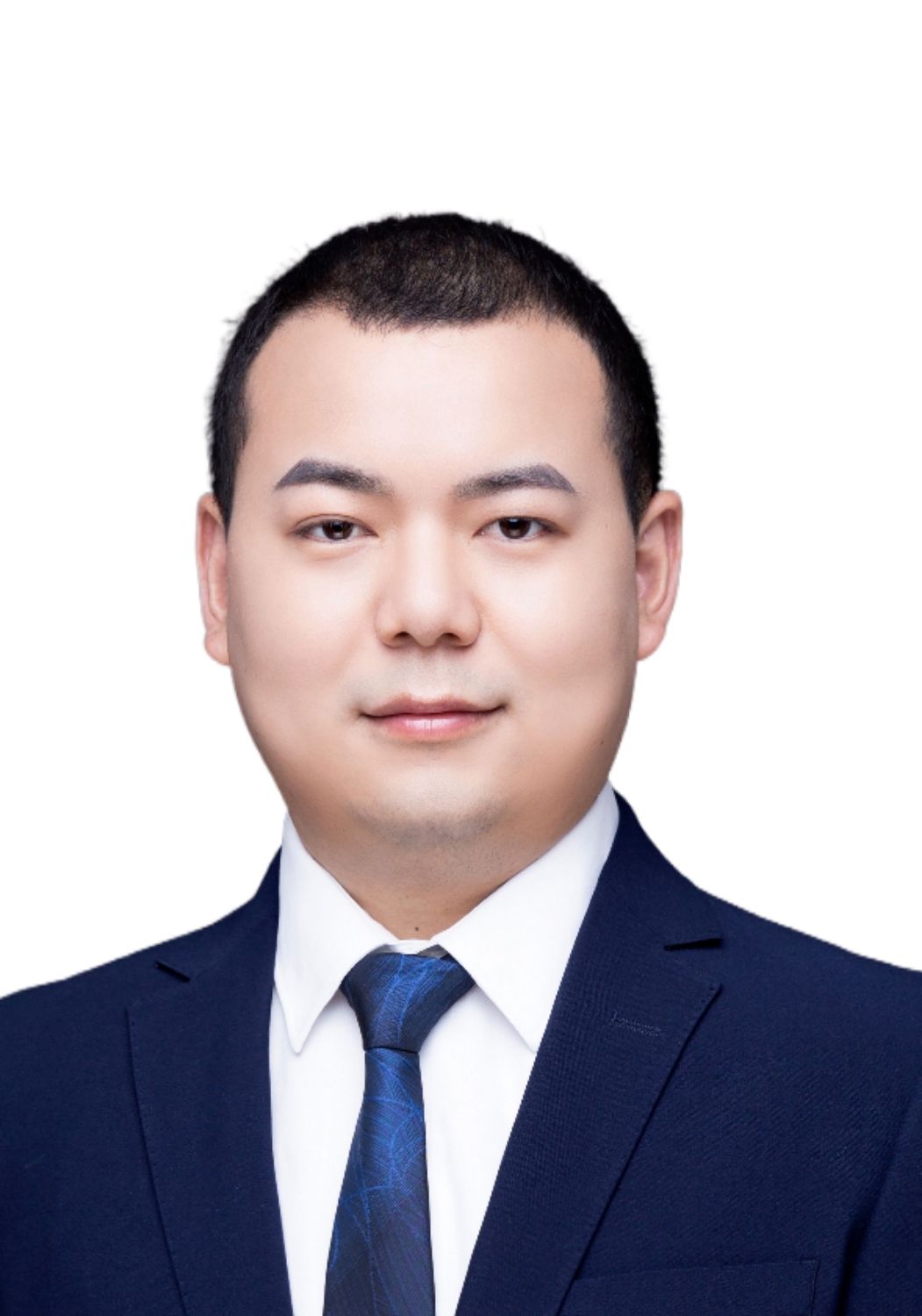}}]{Kai Li}
    is currently an associate professor at Institute of Automation, Chinese Academy of Sciences. He received his Ph.D. degree in pattern recognition and intelligent system from Institute of Automation, Chinese Academy of Sciences in 2018. His main research interest are large-scale imperfect-information games and deep multi-agent reinforcement learning.
\end{IEEEbiography}

\begin{IEEEbiography}[{\includegraphics[width=1in,height=1.25in,clip,keepaspectratio]{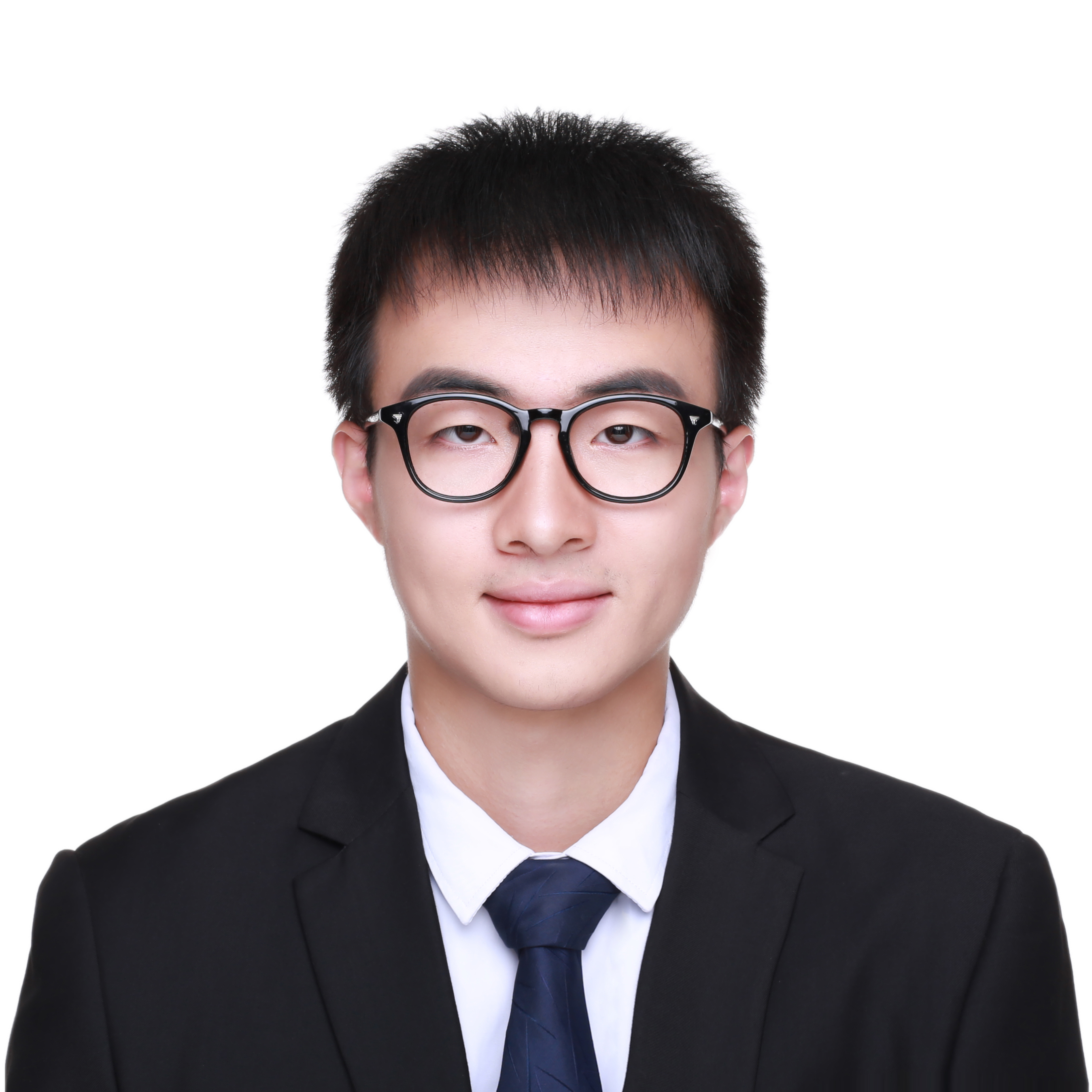}}]{Hang Xu}
	is currently a Ph.D. candidate in pattern recognition and intelligent systems from Institute of Automation, Chinese Academy of Sciences. He received his bachelor's degree in engineering from Wuhan University in 2020. His research interests include computer game and reinforcement learning.
\end{IEEEbiography}

\begin{IEEEbiography}[{\includegraphics[width=1in,height=1.25in,clip,keepaspectratio]{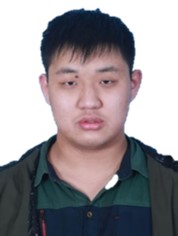}}]{Enmin Zhao}
	 is currently a Ph.D. candidate in pattern recognition and intelligent systems from Institute of Automation, Chinese Academy of Sciences. He received his bachelor's degree in engineering from Tsinghua University in 2018. His research interests include computer poker and deep reinforcement learning.
\end{IEEEbiography}

\begin{IEEEbiography}[{\includegraphics[width=1in,height=1.25in,clip,keepaspectratio]{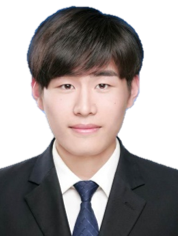}}]{Zhe Wu}
	is currently a master candidate in pattern recognition and intelligent systems from Institute of Automation, Chinese Academy of Sciences. He received his bachelor's degree in engineering from Shandong University in 2019. His research interests include opponent modeling and meta learning.
\end{IEEEbiography}

\begin{IEEEbiography}[{\includegraphics[width=1in,height=1.25in,clip,keepaspectratio]{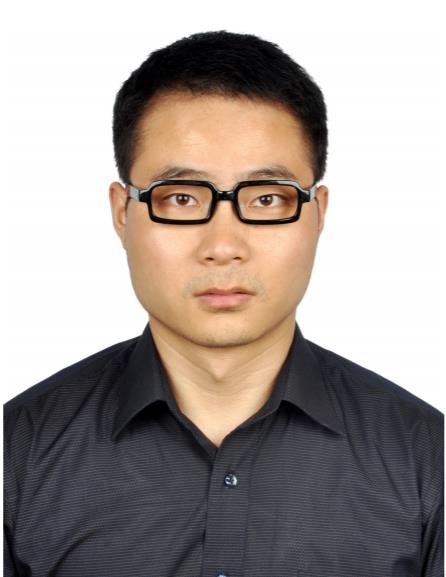}}]{Junliang Xing}
	received his dual B.S. degrees in computer science and mathematics from Xi'an Jiaotong University, Shaanxi, China, in 2007, and the Ph.D. degree in computer science from Tsinghua University, Beijing, China, in 2012. He is currently a Professor with the Institute of Automation, Chinese Academy of Sciences, Beijing, China. His research interests mainly focus on computer vision problems related to human faces and computer gaming problems in imperfect information decision.
\end{IEEEbiography}

\end{document}